\documentclass[letterpaper,twocolumn,10pt]{article}
\usepackage{usenix2019_v3}

\usepackage{tikz}
\usepackage{amsmath}
\usepackage[small,compact]{titlesec}

\usepackage{filecontents}
\usepackage{xspace}
\usepackage{booktabs} %
\usepackage{makecell}
\usepackage{url}
\usepackage{subfig}
\usepackage{ulem} %
\usepackage[export]{adjustbox} %
\usepackage{multirow}
\usepackage{enumitem}
\usepackage{placeins}
\usepackage{cleveref}
\usepackage{tablefootnote} %
\crefformat{figure}{Figure~#2#1#3}
\Crefformat{figure}{Figure~#2#1#3}
\crefformat{table}{Table~#2#1#3}
\Crefformat{table}{Table~#2#1#3}
\crefformat{section}{Section~#2#1#3}
\Crefformat{section}{Section~#2#1#3}

\newcommand{\new}[1]{#1}

\renewcommand{\emph}[1]{\textit{#1}}

\newcommand{\sys}{TraceLab\xspace}

\newcommand{\code}[1]{\ifmmode\text{\texttt{#1}}\else\texttt{#1}\fi}

\newcommand{\prefixtoks}{prefix tokens\xspace}
\newcommand{\Prefixtoks}{Prefix tokens\xspace}      %
\newcommand{\appendtoks}{append tokens\xspace}
\newcommand{\outputtoks}{output tokens\xspace}
\newcommand{\Outputtoks}{Output tokens\xspace}      %
\newcommand{\pfc}{prefix cache\xspace}
\newcommand{\Pfc}{Prefix cache\xspace}              %

\newcommand{\avgstepperrequest}{8.8\xspace}              %
\newcommand{\avgtollcallsperrequest}{10.8\xspace}        %
\newcommand{\avgtimeperrequest}{4.3\xspace}              %
\newcommand{\pntimeperrequest}{6.4\xspace}               %
\newcommand{\mediancachedinputtokens}{119K\xspace}       %
\newcommand{\medianuncachedinputtokens}{875\xspace}      %
\newcommand{\medianoutputtokens}{214\xspace}             %
\newcommand{\mediandecodespeedclaude}{46.8\xspace}             %
\newcommand{\mediandecodespeedcodex}{33.9\xspace}             %
\newcommand{\mediancodecdecodespeed}{61.3\xspace}        %
\newcommand{\mediancodecdecodespeedttft}{3.1\xspace}     %
\newcommand{\totaltoolcatetory}{80\xspace}               %
\newcommand{\toolcallkindsclaude}{54\xspace}             %
\newcommand{\toolcallkindscodex}{31\xspace}              %
\newcommand{\topthreetoolpercent}{80\%\xspace}           %
\newcommand{\toolcalltoppercentage}{80\%\xspace}         %
\newcommand{\toolcalltoppercentagecodex}{95\%\xspace}    %
\newcommand{\toolcallslongerthanonemin}{4\xspace}        %
\newcommand{\toolcallslongerthanoneminpercent}{85\%\xspace} %
\newcommand{\prefixcachehitrate}{95.7\%\xspace}          %
\newcommand{\prefillamplificationfactor}{3.8$\times$\xspace} %

\begin{document}

\date{}

\title{\Large \bf \sys{}: Characterizing Coding Agent Workloads for LLM Serving}

\author{
    {\rm Kan Zhu\textsuperscript{1}} \quad
    {\rm Mathew Jacob\textsuperscript{1}} \quad
    {\rm Chenxi Ma\textsuperscript{1,2,*}} \quad
    {\rm Yi Pan\textsuperscript{3}}\quad
    {\rm Stephanie Wang\textsuperscript{1}}\\
    {\rm Arvind Krishnamurthy\textsuperscript{1}} \quad
    {\rm Baris Kasikci\textsuperscript{1}}\\[0.4em]
    \textsuperscript{1}University of Washington \quad
    \textsuperscript{2}Wuhan University of Technology \quad
    \textsuperscript{3}Shanghai Jiao Tong University
}

\maketitle
\begingroup
\renewcommand{\thefootnote}{*}
\NoHyper
\footnotetext{Work done while interning at the University of Washington.}
\endNoHyper
\endgroup

\begin{abstract}
Coding agents are rapidly becoming a major application of agentic LLMs, but serving them efficiently remains challenging. Progress on this challenge requires understanding real workload patterns, yet the data needed for such analysis is largely absent. Existing public traces and benchmarks do not capture real, day-to-day coding-agent usage across multiple agents and model families for serving-system analysis. To help fill this gap, we collect and release a trace of roughly 4,300 coding-agent sessions, containing about 350,000 LLM steps and 430,000 tool calls from our own day-to-day use of Claude Code and Codex. Our analysis shows that coding-agent workloads feature long autonomous loops, long contexts with short outputs, diverse and heavily-tailed tool calls, and high but imperfect \pfc hit rates. These findings point to concrete opportunities for optimizing serving, including lower-overhead tool calling, append-length-aware prefill, semantic-aware tool-latency prediction, and improved KV-cache management around human-paced gaps. We release the dataset, trace collection pipeline, and analysis code at \url{https://github.com/uw-syfi/TraceLab.git}; the project website is \url{https://tracelab.cs.washington.edu}.

\end{abstract}

\section{Introduction}
\label{sec:intro}

Agentic LLMs have grown rapidly, enabling models to handle increasingly complex tasks through reasoning and tool calling~\cite{yao2023reactsynergizingreasoningacting,schick2023toolformerlanguagemodelsteach}.
Among these, coding agents are evolving especially fast, as companies release their own coding agents, e.g., Claude Code from Anthropic~\cite{claudecode}, Codex from OpenAI~\cite{openaicodex}, and Gemini CLI from Google~\cite{geminicli}. Open-source solutions such as OpenCode~\cite{opencode} and DeepCode~\cite{deepcode}, which allow users to run local model deployments, are also gaining popularity.

From a systems perspective, however, serving these coding agents is challenging. Contexts grow over sessions; SLOs are tight; and repeated tool calling places pressure on both the tool-serving infrastructure and the underlying LLM-serving engines. Understanding the characteristics of these workloads is a prerequisite for serving them efficiently.

Existing benchmarks and traces only partially address this. Traces obtained from serving systems like Mooncake~\cite{qin2025mooncakekvcachecentricdisaggregatedarchitecture} and Splitwise~\cite{patel2024splitwiseefficientgenerativellm} capture real user interactions, but they are not focused on coding agents and lack multi-step tool-call behavior.  Capability benchmarks such as Terminal-Bench~\cite{merrill2026terminalbenchbenchmarkingagentshard} and SWE-bench~\cite{jimenez2024swebenchlanguagemodelsresolve} measure whether an agent can solve a task. They contain relatively few tasks, each narrowly scoped to a single problem. While they are standard for evaluating model accuracy, they are not designed to capture the diversity of real-world coding-agent usage from a serving system's perspective. 

We aim to help fill this gap. Luckily, coding agents will by default log conversations and tool calls, which serves as a rich source of data for understanding real-world usage. We build a pipeline that extracts the data, normalizes the format, anonymizes it to protect user privacy, and finally runs the analysis to understand the workload characteristics.
In this paper, we share initial insights from what is, to our knowledge, the first large-scale cross-provider trace of real coding-agent usage: about 4,300 sessions comprising roughly 350,000 LLM steps and 430,000 tool calls, collected from 43 developers over roughly eight months and spanning both Claude Code and Codex across more than 20 model versions. From our collected traces, we analyze the implications for LLM serving systems.

\textbf {At the session level, the workload is largely autonomous and consists of multiple LLM invocations and tool calls.}.
To complete a user's task, the agent will, on average, carry out \avgstepperrequest LLM calls and invoke tools \avgtollcallsperrequest times. It takes on average \avgtimeperrequest minutes to complete one request, with long tails whose p90 exceeds \pntimeperrequest minutes. For the majority of iterations, the context grows as more user input, tool results, and LLM generations are added. However, two categories of context reduction are observed: a compaction near the context limit and a micro context reduction unique to Codex that occasionally occurs when the user starts their next request.

Due to the multi-step nature of sessions, to avoid repeatedly prefilling the existing conversation history, modern LLM serving systems widely adopt prefix caching~\cite{zheng2024sglangefficientexecutionstructured}. Prefix caching preserves the KV cache for history tokens, so each step only needs to prefill newly appended tokens. Although in API pricing, reading from the prefix cache is roughly $10\times$ cheaper than performing a fresh prefill~\cite{anthropicpricing,openaiapipricing,openaicodexpricing}, the accumulated history in one session can become long, and the cache must be read at every step. As a result, we observe that prefix-cache reads dominate the overall API cost, under the pricing snapshot in \cref{tab:pricing_snapshot}.

\textbf{Within a step, LLM generation features long context but short output.} Despite our expectation of long generations due to reasoning, the frequently occurring tool calls cut LLM generation into multiple steps, making each step's LLM invocation have shorter output length than traditional reasoning workloads. The median LLM generation workload is about \mediancachedinputtokens \prefixtoks (prefix cache read of history context), \medianuncachedinputtokens \appendtoks (fresh new input tokens from user or tool results), and \medianoutputtokens \outputtoks. \new{Despite the long context, both providers achieve good decode speed.} The normalized decode speed has a median of \mediandecodespeedclaude normalized tokens per second for Claude and \mediandecodespeedcodex for Codex, both with significant variance (CV > 50\%). For Codex, the median pure decode speed is \mediancodecdecodespeed with an estimated TTFT of \mediancodecdecodespeedttft seconds per step. 

\textbf{Additionally, tool calls are diverse but heavily tailed for popularity and latency.} While more than \totaltoolcatetory distinct tools are observed, the distribution is heavily skewed. The top 3 tools---Bash, Read, and Edit for Claude; exec\_command, write\_stdin, and apply\_patch for Codex---account for more than \topthreetoolpercent of all tool calls. The latency of tool calls is also diverse, spanning from milliseconds to hours. Tool calls longer than 1\,min are only \toolcallslongerthanonemin percent of all tool calls, but they account for \toolcallslongerthanoneminpercent of total tool-call time.

\textbf{Finally, the \pfc greatly reduces the cost, but misses are still expensive}. The global \pfc hit rate is \prefixcachehitrate, while most misses occur between the last request's final output and the next request's user input, as the delay due to human reading, thinking, and typing can frequently exceed the \pfc eviction time. Overall, cache misses cause \prefillamplificationfactor more tokens to be prefilled than prefills due to truly unique input tokens.

Given these insights, we highlight the following research directions on optimizing LLM serving for coding agents: 
\begin{enumerate}
    \item The overhead of frequent tool-LLM switch motivates denser or fused tool invocations and lower-overhead tool call approval/runtime paths.
    \item The mixture of short and long append lengths motivates append-length-aware prefill routing, along with adaptive selection of kernels and serving-engine parallelism.
    \item The large latency variation for each tool type suggests that tool-latency prediction for KV cache eviction policy should consider the semantics of the requested operation and recent latency history, not just the tool name.
    \item The long prefix motivates sparse attention for reducing the decoding cost. \item The frequent idle gaps necessitate cheaper KV cache storage, a better compression algorithm, and wise KV-cache eviction, or prefetching policies around those gaps.
\end{enumerate}

Overall, our work provides the first large-scale, cross-provider look into the real-world workload characteristics of coding agents. We release our datasets, trace collection and analysis pipeline, and we are excited to see how the community can use these resources to further understand and optimize for this emerging workload.

\section{Background and Motivation}
\label{sec:background}

\subsection{Agentic LLMs and Coding Agents}
Agentic LLMs, compared to traditional LLMs, are designed to interact with external tools and environments, perform multi-step reasoning, and thus are more suitable for complex tasks. Coding agents, as a specific type of agentic LLMs, are designed to assist with programming tasks, such as code generation, debugging, and testing.
In just a few years, they have become one of the most widely adopted applications of agentic LLMs.
The 2025 Stack Overflow Developer Survey reports that 84\% of developers now use or plan to use AI tools in their workflow~\cite{stackoverflow2025}, and individual products have reached enormous scale: GitHub Copilot surpassed 20 million users in 2025~\cite{copilot20m}, Cursor reports more than 7 million monthly active users~\cite{cursorstats}, and Anthropic's Claude Code reached a \$1\,billion annualized revenue run-rate within roughly six months of its public launch~\cite{claudecode2025}.

Agentic programming can be organized as a three-level hierarchy consisting of \emph{sessions}, \emph{requests}, and \emph{steps}.
A \emph{session} is the highest-level unit of interaction.
It maintains the agent's accumulated context, including the conversation history, the contents of files the agent has read or modified, and the outputs of tool calls.
This context persists across user interactions until the session terminates.

Within a session, the user issues one or more \emph{requests}.
Each request corresponds to a natural-language task, such as repairing a failing test or adding a command-line option.
Requests are processed sequentially, and each request inherits the context accumulated by prior requests in the same session.

A request is typically resolved through an \emph{agentic loop} composed of multiple \emph{steps}.
Each step consists of one LLM invocation together with any tool calls it produces.
Given the current context, the LLM reasons about the task, may emit intermediate output, and then either invokes one or more tools or returns a final response to the user.
When tools are invoked, their results are appended to the context and used as input to the next step.
Thus, the first step of a request is \emph{user-initiated}, while subsequent steps are \emph{tool-initiated}.
This loop continues until a step returns a final answer, at which point control returns to the user and the next request may begin from the updated session context.
Concretely, the agentic loop has the following shape:
\[
\begin{array}{r@{\;}c@{\;}l@{\;}c@{\;}l}
\text{User Input}  & \rightarrow & \text{LLM} & \rightarrow & \text{Tool Calls}\\
\text{Tool Results} & \rightarrow & \text{LLM} & \rightarrow & \text{Tool Calls}\\
                   & \vdots     &            &             & \\
\text{Tool Results} & \rightarrow & \text{LLM} & \rightarrow & \text{Final Answer}
\end{array}
\]

\subsection{Existing Datasets and Benchmarks}
Despite the complexity and popularity of coding agents, there is a lack of public datasets that capture their real-world usage and suitable for guiding system optimization.

Widely used workload traces for studying LLM serving systems, such as Mooncake~\cite{qin2025mooncakekvcachecentricdisaggregatedarchitecture}, LMSYS-Chat-1M~\cite{zheng2024lmsyschat1mlargescalerealworldllm}, BurstGPT~\cite{wang2025burstgptrealworldworkloaddataset}, and Splitwise~\cite{patel2024splitwiseefficientgenerativellm}, capture LLM serving traffic at scale, but primarily model traditional usage---human chat, single-turn completion, or short multi-turn interactions---and thus lack the long-horizon, multi-step, tool-calling structure of agentic coding workloads. As a result, they are insufficient for characterizing the workload of coding agents or for evaluating modern serving systems under coding-agent traffic.

\begin{table*}[t]
\centering
\caption{Summary of the collected coding-agent trace. Token counts are in billions (B) and millions (M); the price column is an estimated API list-price equivalent.}
\label{tab:trace_facts}
\small
\setlength{\tabcolsep}{8pt}
\renewcommand{\arraystretch}{1.15}
\begin{tabular}{l r r r}
\toprule
\textbf{Metric} & \textbf{Claude} & \textbf{Codex} & \textbf{Total} \\
\midrule
\multicolumn{4}{@{}l}{\emph{Models}} \\
\quad Model mix (\% within provider)
  & \makecell[tr]{opus-4-7 63.1\%\\opus-4-6 12.0\%\\haiku-4-5 7.9\%\\sonnet-4-6 7.8\%\\opus-4-8 7.7\%\\Other 1.5\%}
  & \makecell[tr]{gpt-5.5 47.5\%\\gpt-5.4 26.0\%\\gpt-5.3-codex 13.7\%\\gpt-5.2-codex 3.4\%\\Other 9.3\%}
  & 23 models \\
\addlinespace
\multicolumn{4}{@{}l}{\emph{Scope \& activity}} \\
\quad Sessions               & 2,676 & 1,589 & 4,265 \\
\quad Distinct users         & 37 & 22 & 43 \\
\quad Observation window     & Oct 2025--Jun 2026 & Sep 2025--Jun 2026 & Sep 2025--Jun 2026 \\
\quad LLM steps             & 140,338 & 216,823 & 357,161 \\
\quad Tool calls             & 142,388 & 290,122 & 432,510 \\
\quad Est.\ API price equiv. & $\sim$\$22.7K & $\sim$\$17.8K & $\sim$\$40.4K \\
\addlinespace
\multicolumn{4}{@{}l}{\emph{Tokens}} \\
\quad Total input tokens               & 28.47\,B & 26.43\,B & 54.90\,B \\
\quad Append tokens                    & 1.19\,B & 1.15\,B & 2.34\,B \\
\quad Prefix tokens                    & 27.28\,B & 25.29\,B & 52.56\,B \\
\quad Output tokens (incl.\ reasoning) & 96.9\,M & 90.1\,M & 186.9\,M \\
\quad Reasoning tokens                 & --- (not reported) & 36.8\,M & --- \\
\bottomrule
\end{tabular}
\end{table*}

Existing benchmarks for coding agents, such as Terminal-Bench~\cite{merrill2026terminalbenchbenchmarkingagentshard} and SWE-bench~\cite{jimenez2024swebenchlanguagemodelsresolve}, consist of realistic programming tasks but are built for a different goal: evaluating model accuracy on isolated tasks. For example, one Terminal-Bench task asks the agent to ``implement an adaptive-rejection sampler,'' a well-scoped problem. In practice, however, such a task is only a single \emph{request} among many in the course of building a real project; replaying these benchmarks therefore captures one request in isolation and underestimates the context growth that accumulates over a session. \new{Benchmark tool usage can also deviate from real-world behavior and lacks standardization: some benchmarks restrict the available tools, such as the bash-only setting in SWE-bench, while others leave the permitted toolset unspecified.} Because the choice of available tools strongly shapes the tool-call distribution, this can materially change the resulting workload characteristics. Finally, although a few thousand tasks suffice for accuracy evaluation, this scale is too small for modeling the behavior of large serving fleets, particularly those using prefill--decode (PD) or attention--FFN (AF) disaggregation.

In short, neither existing datasets nor benchmarks capture coding agents as \emph{persistent, tool-using systems}---the long sessions, repeated tool calls, accumulated context, and human-paced gaps between requests that ultimately shape serving cost. To close this gap, we collect and analyze a large cross-provider trace of real coding-agent usage spanning many developers, agents, and models over an extended period. \new{Because our dataset is collected from daily research workflows that use coding agents for system building, evaluation, and analysis, it directly captures coding agent behaviors in production-like settings, providing valuable insights for understanding coding-agent workloads and informing the design of future LLM serving systems.}

\section{Trace Format and Data Collection}
\label{sec:trace}

\subsection{Raw Agent Logs}
By default, both Claude Code and Codex persist a log for each session that records the full interaction between the user and the agent. Although the two systems use different raw schemas, both expose the same logical event types:
\begin{itemize}
    \item \textbf{Metadata}: session-level information such as the model, permission mode, and title.
    \item \textbf{User messages}: user inputs, including the raw input text.
    \item \textbf{Reasoning}: the agent's internal reasoning steps. The reasoning text is encrypted for both Claude and Codex.
    \item \textbf{Output}: intermediate natural-language output shown to the user, as raw text.
    \item \textbf{Tool calls}: the tools invoked by the agent, with the tool type and raw arguments.
    \item \textbf{Tool results}: the outputs returned by the tools, as raw text.
\end{itemize}
Every event is timestamped, and each LLM invocation reports its token usage.

\subsection{Normalization}
While Claude and Codex logs describe similar interactions, they differ in event structure, token accounting, and tool-timing metadata. To facilitate analysis, we normalize both into a unified, step-level schema in which each row corresponds to a single step, i.e., one LLM invocation together with the tool calls it produces.
\begingroup
\renewcommand{\thefootnote}{\ensuremath{\dagger}}%
\begin{table}[t]
\centering
\caption{Per-session, per-request, and per-step count distributions across the coding-agent trace.}
\label{tab:session_internal_counts}
\small
\setlength{\tabcolsep}{6pt}
\renewcommand{\arraystretch}{1.15}
\begin{tabular}{l r r r r r}
\toprule
\textbf{Metric} & \textbf{Avg} & \textbf{P25} & \textbf{P50} & \textbf{P90} & \textbf{P99} \\
\midrule
\multicolumn{6}{@{}l}{\emph{Per session}} \\
\quad Requests & 9.2 & 1 & 1 & 18 & 137 \\
\quad User-initiated steps\tablefootnote{A \emph{user-initiated step} is a step whose first input event is a user message. This need not equal the number of user messages: (i)~a message the user sends while the agent is still working is delivered together with the next tool result, so that step is counted as tool-initiated; and (ii)~a message that never triggers another model call is not counted as a step at all.} & 8.9 & 1 & 1 & 17 & 129 \\
\quad Tool-initiated steps & 73.6 & 4 & 15 & 135 & 1,107 \\
\quad Tool calls & 101.4 & 8 & 25 & 176 & 1,438 \\
\addlinespace
\multicolumn{6}{@{}l}{\emph{Per request}} \\
\quad User-initiated steps & 0.9 & 1 & 1 & 1 & 1 \\
\quad Tool-initiated steps & 7.8 & 0 & 1 & 20 & 86 \\
\quad Tool calls & 10.8 & 0 & 2 & 30 & 113 \\
\addlinespace
\multicolumn{6}{@{}l}{\emph{Per step}} \\
\quad Tool calls & 1.2 & 1 & 1 & 2 & 4 \\
\bottomrule
\end{tabular}
\end{table}
\endgroup

\noindent\textbf{Token accounting.}
\new{From a serving perspective, we decompose the total input tokens of each step, $I$, into two components: \prefixtoks $P$, the history tokens retrieved from the prefix cache, and \appendtoks $A$, the tokens appended to the context in the current iteration, including user messages, tool results, and tokens introduced by prefix-cache misses.}
The providers report these components as follows:
\[
\begin{aligned}
P_{\mathrm{Claude}} &= \code{cache\_read\_input\_tokens},\\
A_{\mathrm{Claude}} &= \code{input\_tokens}\\
&\quad + \code{cache\_creation\_input\_tokens},\\
P_{\mathrm{Codex}} &= \code{cached\_input\_tokens},\\
A_{\mathrm{Codex}} &= \code{input\_tokens\_total}\\
&\quad - \code{cached\_input\_tokens}.
\end{aligned}
\]
 
We directly use \code{output\_tokens} ($O$) reported by Claude and Codex for total generation length, which already includes reasoning tokens. Codex additionally reports separated \code{reasoning\_output\_tokens}, which we use later to characterize Codex's generation timing. 

Throughout the paper we refer to these three categories as \prefixtoks ($P$), \appendtoks ($A$), and \outputtoks ($O$).

\noindent\textbf{Timing.}
Each step also retains an ordered list of timing events---user messages, tool results, reasoning, output text, and tool calls---from which we reconstruct its internal timeline. For tool latency, Codex reports the tool execution time explicitly, \new{whereas for Claude we derive it from the tool's call and return timestamps.}

\subsection{Anonymization}
To protect user privacy, we release only a sanitized trace. We replace all session, tool-call, project, and user identifiers with stable pseudonyms, and drop the raw user messages and tool input/output text, keeping only their character counts along with token usage, timestamps, and pseudonymous user identifiers for analysis. We summarize the resulting trace in \cref{tab:trace_facts}.

\begin{table}[t]
\centering
\caption{Same-session context change, by step trigger. Each value is the per-step change in total input length (prefix tokens $+$ append tokens) versus the previous step in the session.}
\label{tab:context_growth_and_compaction}
\small
\setlength{\tabcolsep}{6pt}
\renewcommand{\arraystretch}{1.15}
\begin{tabular}{l r r}
\toprule
\textbf{Metric} & \textbf{Claude} & \textbf{Codex} \\
\midrule
\multicolumn{3}{@{}l}{\emph{All steps}} \\
\quad Steps & 137,629 & 210,221 \\
\quad Positive (growth) \% & 99.60\% & 96.56\% \\
\quad Negative (reduction) \% & 0.39\% & 3.43\% \\
\quad\quad Micro \% & 0.06\% & 1.08\% \\
\quad\quad Ordinary \% & 0.09\% & 1.73\% \\
\quad\quad Major reduction \% & 0.24\% & 0.62\% \\
\quad Avg positive growth & 1,719 & 1,838 \\
\addlinespace
\multicolumn{3}{@{}l}{\emph{User-initiated steps}} \\
\quad Steps & 16,927 & 17,033 \\
\quad Positive (growth) \% & 98.21\% & 68.63\% \\
\quad Negative (reduction) \% & 1.71\% & 31.32\% \\
\quad\quad Micro \% & 0.37\% & 12.47\% \\
\quad\quad Ordinary \% & 0.57\% & 18.09\% \\
\quad\quad Major reduction \% & 0.77\% & 0.76\% \\
\quad Avg positive growth & 1,527 & 2,742 \\
\addlinespace
\multicolumn{3}{@{}l}{\emph{Tool-initiated steps}} \\
\quad Steps & 120,702 & 193,188 \\
\quad Positive (growth) \% & 99.79\% & 99.03\% \\
\quad Negative (reduction) \% & 0.21\% & 0.97\% \\
\quad\quad Micro \% & 0.02\% & 0.07\% \\
\quad\quad Ordinary \% & 0.03\% & 0.29\% \\
\quad\quad Major reduction \% & 0.16\% & 0.61\% \\
\quad Avg positive growth & 1,746 & 1,783 \\
\bottomrule
\end{tabular}
\end{table}

\section{Session and Context Management}
In this section, we analyze coding-agent macroscopic metrics, including the agentic loop, the context management, and the cost and wall-clock time each session consumes.

\subsection{Autonomous Loop}
\label{sec:session_loop}
First, we characterize the distribution of the number of requests, tool calls, and steps in \cref{tab:session_internal_counts} . A session on average has more than 9 requests, with a p99 of 137 requests, which highlights its persistent nature. Most of the steps are tool-initiated, so the loop is mostly autonomous. To solve one request, an agent on average takes around 8 steps with 11 tool calls. Each step on average uses slightly more than 1 tool call, suggesting that some tool calling is parallelized, but this phenomenon is not pervasive.

\subsection{Context Growth and Compaction}

Next we investigate the context-management of coding agents. We show the statistics in \cref{tab:context_growth_and_compaction}. In most cases, context accumulates and positive growth happens, since new user input, tool results, and output all add to the prior context. However, we do observe negative growth. We categorize the reduction into micro-reduction (0--1024 tokens), ordinary reduction (1k--64k tokens), and major reduction (more than 64k tokens). For Claude, reduction overall is very rare, and mostly major reduction. Codex, however, has more frequent reductions, mostly occurring at user-initiated steps, and most of them are micro to ordinary reductions.

One key type of context reduction is \emph{context compaction}, triggered when the context nears the model's limit: prior context is summarized and the session starts accumulating again from a very short history. We distinguish it from an arbitrary major reduction with three criteria: the total input must (i)~drop by at least 64k tokens in one step, (ii)~do so while near the session's peak context (the pre-drop level is at least 75\% of the session's observed maximum), and (iii)~recover slowly, never rebounding to 75\% of the pre-drop level within the next three steps. \cref{tab:session_compaction} reports the result. Most major reductions are genuine compactions (1,519 of 1,630). 

Compaction is not rare but not dominant either: 9.7\% of sessions undergo at least one, and those that do average 3.7 with a long tail (p99 of 33). It is overwhelmingly tool-initiated (86.5\%, i.e.\ occurring mid-loop) rather than user-initiated, and far more common in Codex (18.4\% of sessions) than Claude (4.5\%), consistent with Codex's shorter context length.

\begin{table}[t]
\centering
\caption{Context compactions per session: a near-limit total-input drop ($\geq$64k) that recovers slowly (no rebound to 75\% of the pre-drop level within three steps).}
\label{tab:session_compaction}
\small
\setlength{\tabcolsep}{5pt}
\renewcommand{\arraystretch}{1.15}
\begin{tabular}{l r r}
\toprule
\textbf{Metric} & \textbf{Claude} & \textbf{Codex} \\
\midrule
Sessions & 2,676 & 1,589 \\
Major reductions ($\geq$64k drop) & 324 & 1,306 \\
\quad of which compactions & 284 (87.7\%) & 1,235 (94.6\%) \\
Sessions with $\geq$1 compaction & 120 (4.5\%) & 292 (18.4\%) \\
\addlinespace
\multicolumn{3}{@{}l}{\emph{Compactions per session}} \\
\quad Avg (all sessions) & 0.106 & 0.777 \\
\quad Avg (sessions with $\geq$1) & 2.37 & 4.23 \\
\quad P90 / P99 (sessions with $\geq$1) & 6 / 12 & 8 / 34 \\
\addlinespace
\multicolumn{3}{@{}l}{\emph{Trigger}} \\
\quad User-initiated & 105 (37.0\%) & 100 (8.1\%) \\
\quad Tool-initiated & 179 (63.0\%) & 1,135 (91.9\%) \\
\bottomrule
\end{tabular}
\end{table}

\begin{table}[t]
\centering
\caption{Per-session, per-request, and per-step cost (USD) by category.}
\label{tab:cost_distribution}
\small
\setlength{\tabcolsep}{4pt}
\renewcommand{\arraystretch}{1.15}
\begin{tabular}{l r r r r r}
\toprule
\textbf{Metric} & \textbf{Avg} & \textbf{P50} & \textbf{P90} & \textbf{P99} & \textbf{\% cost} \\
\midrule
\multicolumn{6}{@{}l}{\emph{Per session}} \\
\quad Total & \$9.70 & \$0.61 & \$13.4 & \$178 &  \\
\quad Append tokens & \$2.83 & \$0.24 & \$3.62 & \$46.5 & 29.2\% \\
\quad Prefix tokens & \$5.77 & \$0.17 & \$7.55 & \$111 & 59.5\% \\
\quad Output tokens & \$1.09 & \$0.16 & \$1.92 & \$16.9 & 11.2\% \\
\addlinespace
\multicolumn{6}{@{}l}{\emph{Per request}} \\
\quad Total & \$1.01 & \$0.33 & \$2.44 & \$9.33 &  \\
\quad Append tokens & \$0.29 & \$0.04 & \$0.73 & \$3.91 & 29.2\% \\
\quad Prefix tokens & \$0.60 & \$0.16 & \$1.35 & \$6.51 & 59.5\% \\
\quad Output tokens & \$0.11 & \$0.03 & \$0.29 & \$1.08 & 11.2\% \\
\addlinespace
\multicolumn{6}{@{}l}{\emph{Per step}} \\
\quad Total & \$0.11 & \$0.07 & \$0.20 & \$0.72 &  \\
\quad Append tokens & \$0.03 & \$0.00 & \$0.03 & \$0.68 & 29.2\% \\
\quad Prefix tokens & \$0.07 & \$0.05 & \$0.12 & \$0.41 & 59.5\% \\
\quad Output tokens & \$0.01 & \$0.01 & \$0.03 & \$0.12 & 11.2\% \\
\bottomrule
\end{tabular}
\end{table}

\subsection{Cost Distribution}
\label{sec:cost_distribution}

We now compute the price of each session, request, and step. We use the providers' published API pricing for the models~\cite{anthropicpricing,openaiapipricing,openaicodexpricing}, summarized in \cref{tab:pricing_snapshot}, and categorize tokens into \appendtoks, \prefixtoks, and \outputtoks. The result is shown in \cref{tab:cost_distribution}.

\begin{table}[t]
\centering
\caption{API pricing snapshot used for cost accounting. Rates are USD per million tokens from provider pricing pages~\cite{anthropicpricing,openaiapipricing,openaicodexpricing}; Claude cache-write rates use the 5-minute prompt-cache tier.}
\label{tab:pricing_snapshot}
\small
\setlength{\tabcolsep}{3pt}
\renewcommand{\arraystretch}{1.12}
\begin{tabular}{l l r r r r}
\toprule
\textbf{Provider} & \textbf{Model group} & \makecell[r]{\textbf{Input}} & \makecell[r]{\textbf{5m}\\\textbf{write}} & \makecell[r]{\textbf{Cache}\\\textbf{read}} & \textbf{Output} \\
\midrule
Claude & Opus 4.6--4.8 & 5.00 & 6.25 & 0.50 & 25.00 \\
Claude & Sonnet 4.6 & 3.00 & 3.75 & 0.30 & 15.00 \\
Claude & Haiku 4.5 & 1.00 & 1.25 & 0.10 & 5.00 \\
Codex & GPT-5.5 & 5.00 & -- & 0.50 & 30.00 \\
Codex & GPT-5.4 & 2.50 & -- & 0.25 & 15.00 \\
\bottomrule
\end{tabular}
\end{table}

A session costs \$9.70 on average but only \$0.61 at the median, with a heavy tail (p99 of \$178) driven by a few very long sessions; a request costs \$1.01 on average (median \$0.33) and a step \$0.11.

\Prefixtoks dominate spend. Although \prefixtoks are billed at roughly one-tenth the fresh-input rate, their volume---the accumulating context replayed on every step---makes them 59.5\% of total cost, versus 29.2\% for \appendtoks, including Claude 5-minute cache-write charges, and only 11.2\% for \outputtoks. \Outputtoks are cheap in aggregate despite their high per-token price simply because each step emits few tokens (median \medianoutputtokens). This inverts the usual intuition that generation is the expensive part: for coding agents the cost is overwhelmingly in re-reading context.

\subsection{Timing Distribution}

Finally, we break down wall-clock time in \cref{tab:timing_distribution}. We decompose time into three components: \emph{human thinking} (the gap from the previous event to the next user message), \emph{LLM generation} (the observed time for the model to generate reasoning, output, and tool inputs), and \emph{tool execution} (effective tool latency). Because human thinking happens \emph{between} requests, request response time contains only generation and tool execution; the final block reports the individual positive human waits, generation spans, and tool latencies that match the corresponding CDF/summary distributions.

\begin{table}[t]
\centering
\caption{Per-session, per-request, per-step, and individual latency wall-clock time by category.}
\label{tab:timing_distribution}
\small
\setlength{\tabcolsep}{4pt}
\renewcommand{\arraystretch}{1.15}
\begin{tabular}{l r r r r r}
\toprule
\textbf{Metric} & \textbf{Avg} & \textbf{P50} & \textbf{P90} & \textbf{P99} & \textbf{\% time} \\
\midrule
\multicolumn{6}{@{}l}{\emph{Per session}} \\
\quad Total elapsed & 8.2h & 5.1m & 5.9h & 206.5h &  \\
\quad Human thinking & 7.6h & 0.0s & 4.1h & 190.0h & 92.3\% \\
\quad LLM generation & 16.1m & 2.0m & 26.7m & 3.9h & 3.3\% \\
\quad Tool execution & 23.5m & 14.9s & 26.1m & 6.9h & 4.8\% \\
\addlinespace
\multicolumn{6}{@{}l}{\emph{Per session, human capped (1h)}} \\
\quad Total & 1.8h & 5.5m & 3.4h & 27.2h &  \\
\quad Human thinking & 1.2h & 0.0s & 2.3h & 18.3h & 64.3\% \\
\quad LLM generation & 16.1m & 2.0m & 26.7m & 3.9h & 14.5\% \\
\quad Tool execution & 23.5m & 14.9s & 26.1m & 6.9h & 21.2\% \\
\addlinespace
\multicolumn{6}{@{}l}{\emph{Per request}} \\
\quad Total (response time) & 4.3m & 38.3s & 6.4m & 43.9m &  \\
\quad LLM generation & 1.7m & 28.8s & 3.7m & 13.6m & 41.0\% \\
\quad Tool execution & 2.5m & 0.3s & 2.3m & 34.4m & 59.8\% \\
\addlinespace
\multicolumn{6}{@{}l}{\emph{Per step}} \\
\quad LLM generation & 11.5s & 4.9s & 20.2s & 1.3m & 40.7\% \\
\quad Tool execution & 16.8s & 0.1s & 10.0s & 3.2m & 59.3\% \\
\addlinespace
\multicolumn{6}{@{}l}{\emph{Per individual latency}} \\
\quad Human thinking & 46.7m & 1.4m & 20.6m & 13.9h &  \\
\quad LLM generation & 13.2s & 5.7s & 22.2s & 1.4m &  \\
\quad Tool execution & 18.4s & 0.3s & 13.6s & 3.6m &  \\
\bottomrule
\end{tabular}
\end{table}

From the table, the coding session is mostly idle waiting on the human: human thinking is 92.3\% of session wall-clock, surpassing LLM generation (3.3\%) and tool execution (4.8\%). Most sessions are short---the median session is a single request with no inter-request gap---but a heavy tail of sessions left open for hours or days (session p99 elapsed of $\sim$206h) accumulates most of the idle time. In the individual latency view, positive human-input waits have median 1.4m and p90 20.6m; observed LLM-generation spans have median 5.7s and p90 22.2s; and positive tool latencies have median 0.3s and p90 13.6s. When we cap each idle gap at one hour (the \emph{human capped} block), the human share of the now cache-relevant budget falls to 64.3\%, with generation and tool execution rising to 14.5\% and 21.2\%.

Within an individual request, tool execution and generation both contribute to the response time: of the total 2{,}783 hours of response time, tools account for 59.8\% versus 41.0\% for generation. An average request takes 4.3 minutes end to end (median 38s, p90 6.4 min).

\subsection{Takeaways and Systems Opportunities}
\textbf{Takeaways}
\begin{itemize}
    \item \textbf{Autonomous, heavy-tailed sessions.} The loop is largely self-driving---most steps are tool-initiated---and session length is heavy-tailed, with a few very long sessions dominating.
    \item \textbf{Compaction is rare.} Major context reduction is rare; when it happens it is usually a genuine compaction near the context limit, with Codex additionally showing frequent \emph{micro}-reductions that Claude does not.
    \item \textbf{Cost spent on \prefixtoks.} \Prefixtoks, not generation, account for the majority of cost.
    \item \textbf{Human-bound sessions.} Idle time waiting on the human dominates the session end-to-end time, while within a request tool execution and generation contribute comparable time.
\end{itemize}

\textbf{Systems Opportunities}
\begin{itemize}
    \item \textbf{Denser tool calling.} Tool calls are frequent, yet most steps issue only one. Encouraging more tool calls per step---or \emph{fusing} them---would amortize the LLM$\leftrightarrow$tool scheduling overhead and expose more tool-level parallelism.
    \item \textbf{Lower cache costs.} Because cache reads are a leading cost driver, the prior context is worth storing and reloading more efficiently, including cost-effective KV-cache storage hardware and infrastructure, and sparse-attention schemes that shrink the prefix loading cost.
\end{itemize}

\section{LLM generation}
Next, we focus on the LLM generation and investigate the input, output and timing of the generation. 
\begin{table}[t]
\centering
\caption{Per-step prefix, append, and output token length distribution.}
\label{tab:token_length_distribution}
\small
\setlength{\tabcolsep}{4pt}
\renewcommand{\arraystretch}{1.15}
\begin{tabular}{l r r r r r}
\toprule
\textbf{Tokens} & \textbf{Avg} & \textbf{P25} & \textbf{P50} & \textbf{P90} & \textbf{P99} \\
\midrule
\multicolumn{6}{@{}l}{\emph{Prefix tokens}} \\
\quad Claude & 194,361 & 59,949 & 126,180 & 467,082 & 918,111 \\
\quad Codex & 116,623 & 67,456 & 115,584 & 201,600 & 231,040 \\
\addlinespace
\multicolumn{6}{@{}l}{\emph{Append tokens}} \\
\quad Claude & 8,479 & 384 & 857 & 5,342 & 232,206 \\
\quad Codex & 5,283 & 406 & 886 & 8,310 & 121,009 \\
\addlinespace
\multicolumn{6}{@{}l}{\emph{Output tokens}} \\
\quad Claude & 690 & 132 & 252 & 1,671 & 6,571 \\
\quad Codex & 415 & 70 & 184 & 939 & 3,508 \\
\bottomrule
\end{tabular}
\end{table}

\subsection{Input length distribution}
First, we investigate the per-step input length, split into \prefixtoks (the replayed accumulated context) and \appendtoks (the freshly added, uncached input). \cref{tab:token_length_distribution} shows the distribution for both providers.
A median Claude step reads back 126k \prefixtoks but appends only 857, while Codex reads 116k and appends 886---roughly two orders of magnitude more \prefixtoks than \appendtoks. Because Claude has a longer context length, its prefix stretches to a p99 of 918k tokens, while Codex saturates near 231k.

\begin{figure}[t]
\centering
\includegraphics[width=\columnwidth]{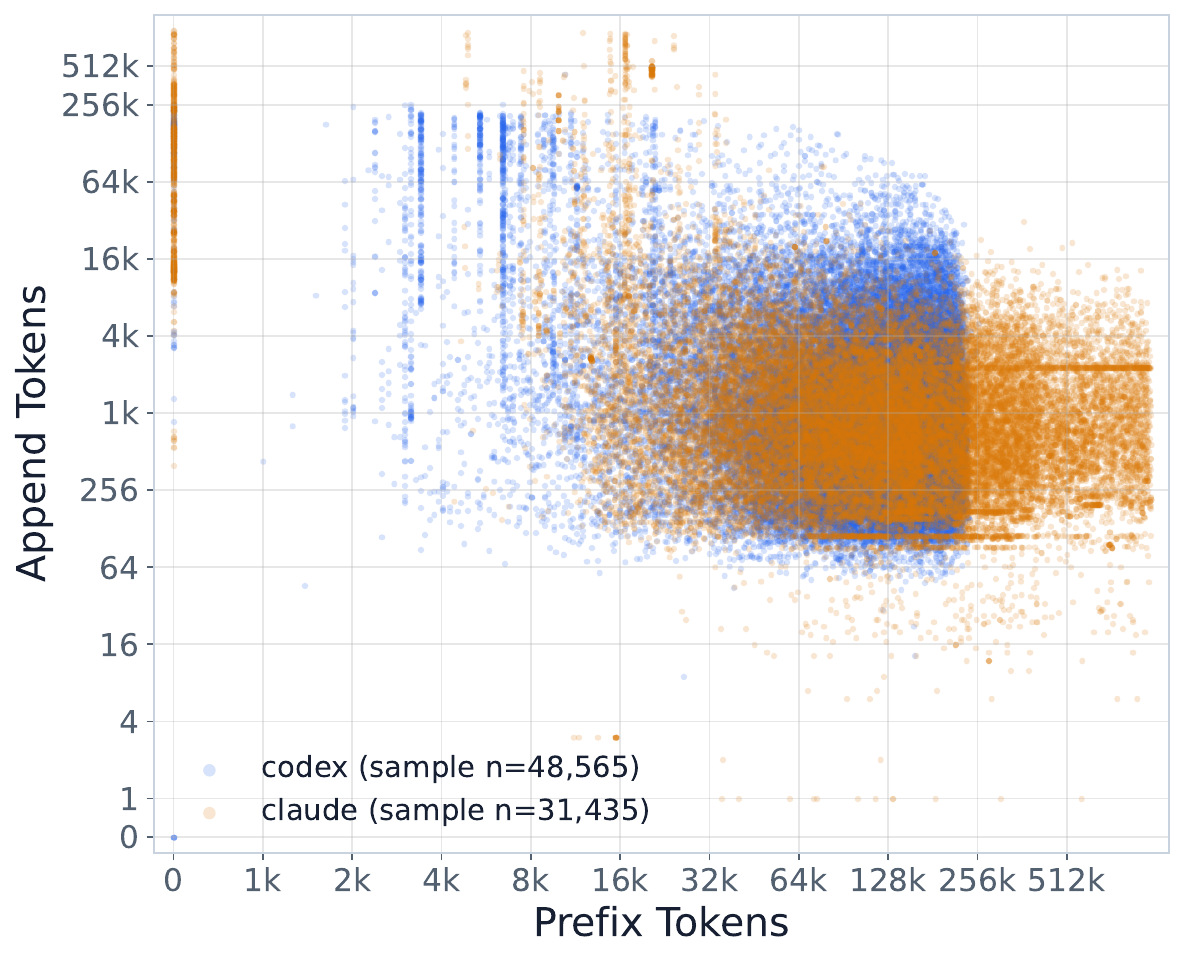}
\caption{Per-step \prefixtoks vs.\ \appendtoks.}
\label{fig:prefill_append_relationship}
\end{figure}

In \cref{fig:prefill_append_relationship}, we further analyze the relationship between prefix and append lengths. Most of the data points fall in prefix length 32k-128k and append 256-8k. We can also see two major groups, one group, with short prefix length (<16k) mainly have relatively large \appendtoks. They mostly represent \pfc miss or initial prefills. The other group, with longer prefix, having much shorter \appendtoks, representing normal context growth due to tool result or user input.

\cref{tab:append_by_prefix} quantifies this split. When there are few \prefixtoks the \appendtoks dominate---at a prefix below 1k the median append is 78k tokens for Claude and 124k for Codex---whereas once the prefix grows past 32k the median collapses to well under 1k, as those steps carry only an incremental tool result or user turn.

\begin{table}[t]
\centering
\caption{Append tokens per step, conditioned on the number of prefix tokens, for each provider.}
\label{tab:append_by_prefix}
\small
\setlength{\tabcolsep}{5pt}
\renewcommand{\arraystretch}{1.15}
\begin{tabular}{l r r r r r}
\toprule
\textbf{Prefix} & \textbf{Steps} & \textbf{Avg} & \textbf{P50} & \textbf{P90} & \textbf{P99} \\
\midrule
\multicolumn{6}{@{}l}{\emph{Claude}} \\
\quad $<$1k & 2,937 & 136.1K & 78.4K & 344.3K & 871.9K \\
\quad 1--2k & 0 & -- & -- & -- & -- \\
\quad 2--4k & 2 & 2.7K & 2.7K & 3.8K & 4.0K \\
\quad 4--8k & 530 & 122.0K & 17.2K & 385.3K & 881.3K \\
\quad 8--16k & 4,034 & 38.8K & 3.5K & 108.5K & 549.1K \\
\quad 16--32k & 10,248 & 35.4K & 1.3K & 30.7K & 661.0K \\
\quad 32--64k & 20,919 & 2.8K & 951 & 5.3K & 27.3K \\
\quad 64--128k & 33,571 & 1.6K & 793 & 3.7K & 12.8K \\
\quad 128--256k & 34,840 & 1.4K & 710 & 3.2K & 10.1K \\
\quad $>$256k & 33,257 & 1.4K & 762 & 3.1K & 8.8K \\
\addlinespace
\multicolumn{6}{@{}l}{\emph{Codex}} \\
\quad $<$1k & 626 & 116.3K & 124.3K & 210.7K & 247.0K \\
\quad 1--2k & 90 & 22.2K & 4.3K & 63.9K & 192.6K \\
\quad 2--4k & 2,108 & 56.4K & 20.8K & 168.7K & 240.8K \\
\quad 4--8k & 3,501 & 60.2K & 25.7K & 172.4K & 220.7K \\
\quad 8--16k & 5,503 & 22.8K & 2.9K & 84.7K & 195.5K \\
\quad 16--32k & 10,470 & 9.6K & 1.9K & 18.8K & 152.2K \\
\quad 32--64k & 29,925 & 3.7K & 954 & 8.3K & 50.4K \\
\quad 64--128k & 72,996 & 2.7K & 796 & 6.1K & 31.3K \\
\quad 128--256k & 91,598 & 2.2K & 771 & 5.3K & 21.0K \\
\quad $>$256k & 6 & 750 & 900 & 1.1K & 1.1K \\
\bottomrule
\end{tabular}
\end{table}

\begin{figure}[t]
\centering
\includegraphics[width=\columnwidth]{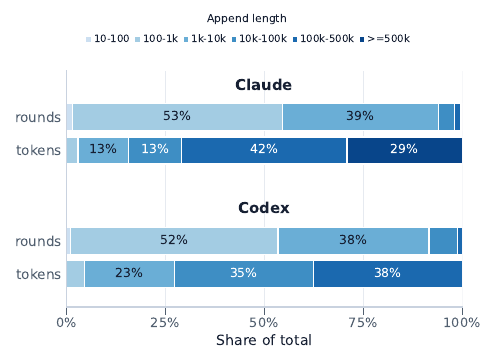}
\caption{Per-step append length by step count and by total append-token share, split by provider.}
\label{fig:prefill_weighted_bar}
\end{figure}

While long-append steps are rare, they contribute the majority of total prefill workload. \cref{fig:prefill_weighted_bar} shows this effect. For each provider, the upper bar is the share of \emph{steps} and the lower bar the share of total \emph{append tokens}, split by per-step append length. Over 90\% of steps append fewer than 1k tokens, yet more than 70\% of all \appendtoks come from the rare steps that append 10k or more.

\subsection{Output length distribution}
In contrast to the large inputs, outputs are short. \cref{tab:token_length_distribution} shows a median of only 252 \outputtoks for Claude and 184 for Codex, with the p90 still under 1.7k. Even at the p99, \outputtoks stay in the few-thousand-token range, far below the input sizes. This is counterintuitive, by can be explained by the frequent tool calls. As we show in \cref{sec:session_loop}, one full response are effectively cut into average 8 tool call steps, thus the individual output length are short, and sometimes only generates next tool call parameters.

\begin{figure}[!t]
\centering
\includegraphics[width=\columnwidth]{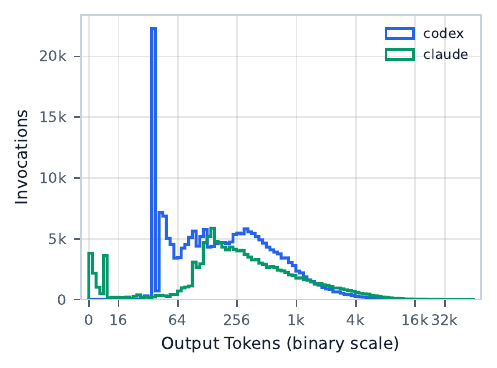}
\caption{Per-step output-token distribution by provider.}
\label{fig:output_tokens}
\end{figure}

\cref{fig:output_tokens} shows the full per-step output distribution: both providers concentrate in the low hundreds of tokens, and Codex additionally has a pronounced spike of very short ($\sim$40-token) generations, due to its popular tool call \texttt{write\_stdin}, that wait for prior command to finish or sending \texttt{Ctrl+C} for interruption.

\begin{figure*}[!t]
\centering
\includegraphics[width=0.85\textwidth]{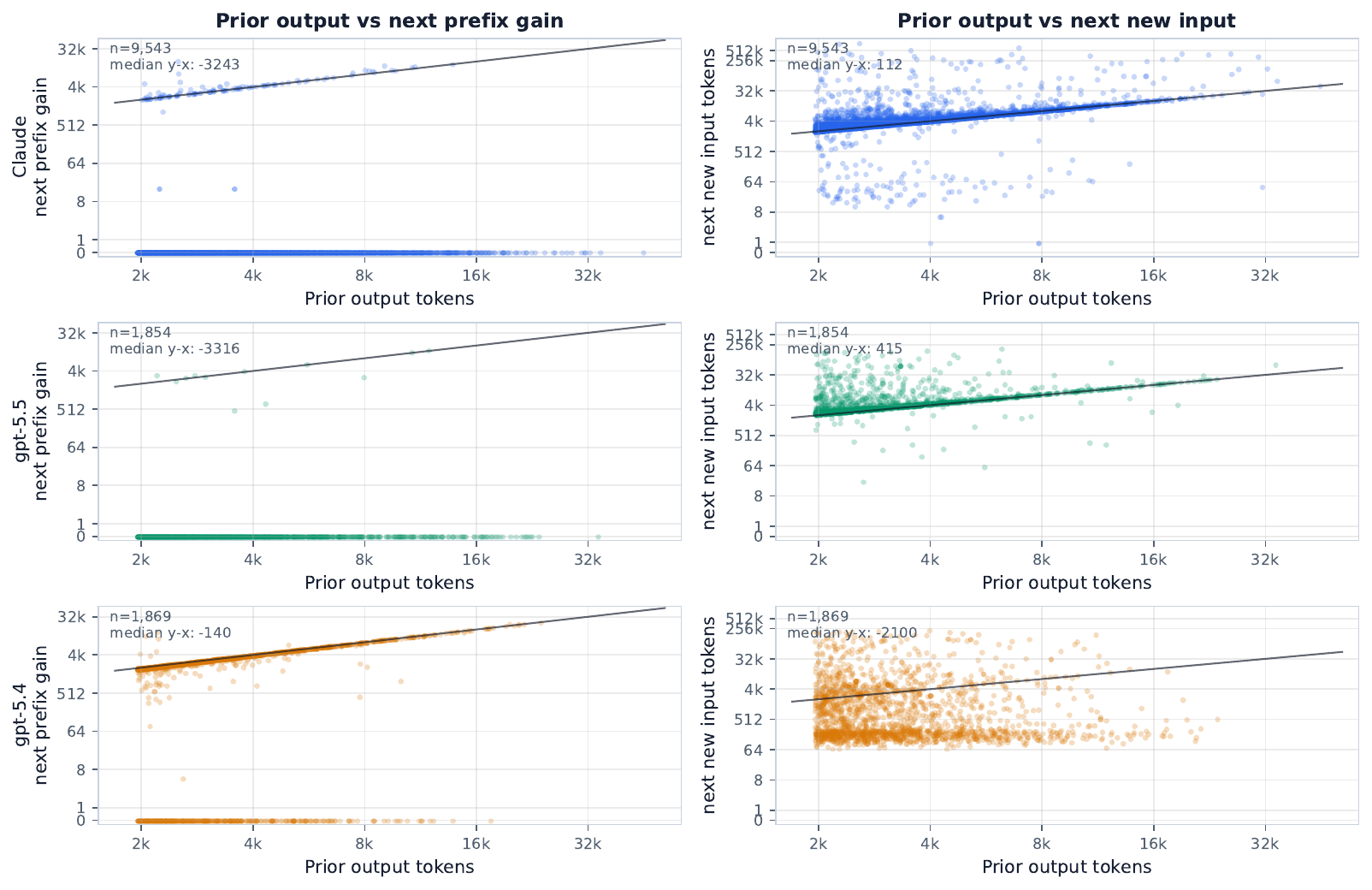}
\caption{Merged output-attribution evidence by model for previous outputs of at least 2k tokens. Left: prior output versus next-step prefix gain; right: prior output versus next-step \appendtoks.}
\label{fig:model_merged_output_attribution}
\end{figure*}

\begin{figure}[!t]
\centering
\includegraphics[width=\columnwidth]{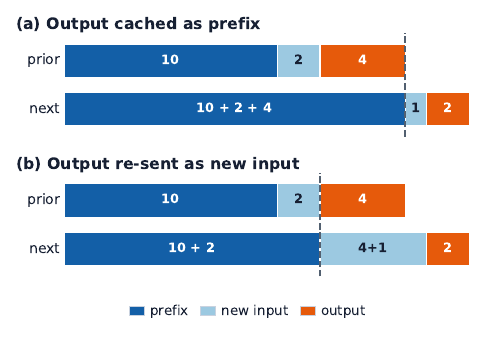}
\caption{Two ways a prior step's output can be accounted in the next step (schematic; bar lengths illustrative).}
\label{fig:output_attribution}
\end{figure}

\subsection{Output token attribution}
Next, we investigate how a prior step's \outputtoks are accounted for in the next step's prompt. If the serving system saves the newly produced KV-cache entries during generation, then the next step can reuse the prior step's whole prompt composition: \prefixtoks, \appendtoks, and \outputtoks. We call this case \emph{output-cached}. However, we also observe cases where the next step's prefix excludes the prior output and instead re-sends it as part of the next step's \appendtoks. We call this case \emph{output-resend}.

\Cref{fig:output_attribution} shows the two accounting schemes schematically. In \cref{fig:output_attribution}(a), the prior step has 10 units of prefix, 2 units of append, and 4 units of output; all 16 units become the next step's prefix, followed by that step's own append and output. In \cref{fig:output_attribution}(b), only the prior prefix and append are cached ($10+2=12$). The prior output is re-sent as part of the next step's append; with one additional unit of fresh input, the next append becomes $4+1=5$.

We distinguish the two cases using their key invariants. Under output-cached, the next step's prefix gain---the next prefix minus the prior prefix and append---should track the prior output. Under output-resend, the next prefix should remain close to the sum of the prior prefix and append, while the next step's append should include the prior output. \cref{fig:model_merged_output_attribution} applies this test separately within each model and focuses on prior outputs of at least 2k tokens, where the signal is large enough to be visible. The figure indicates that Claude primarily uses output-resend. Codex, however, changes behavior across versions: \texttt{gpt-5.4} is mostly output-cached, while \texttt{gpt-5.5} is mostly output-resend.

We hypothesize that this difference reflects KV-cache pool management choices in PD-disaggregated serving. The output-cached policy requires transferring the KV entries produced during decode back to the shared KV-storage backend. The next prefill instance can then load those entries, extend the cache for the next prompt, write the updated KV entries back to storage, and hand them off to the decode instance. This path is more complex, but it avoids re-prefilling the prior output. The output-resend policy, in contrast, only requires prefill instances to write to the shared KV-storage backend. Decode instances are read-only with respect to shared KV state: they fetch the prefix KV needed for generation, but discard the KV entries for newly generated \outputtoks.

\subsection{Timing}
\begin{figure}[!t]
\centering
\includegraphics[width=\columnwidth]{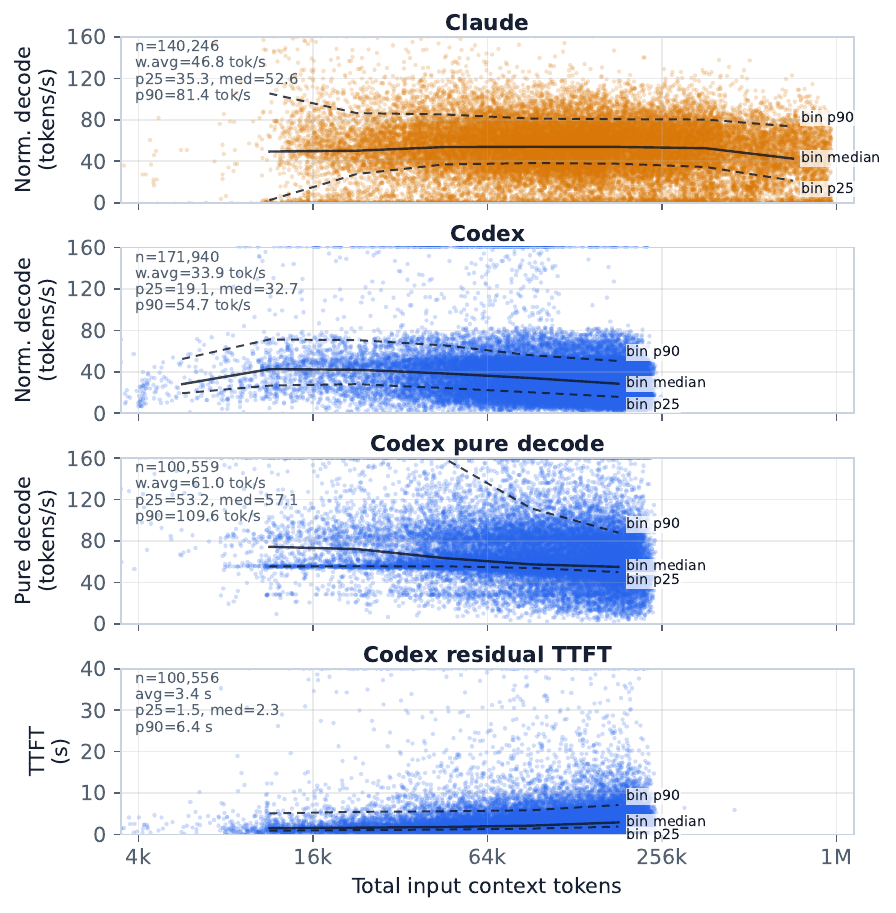}
\caption{Trace-observed LLM timing versus total input context length. }
\label{fig:context_decode_speed}
\end{figure}

We next examine trace-observed LLM generation latency. Let \(t_{\mathrm{input}}\) be the latest input-event timestamp (\texttt{user\_message} or \texttt{tool\_result}) before the first model output, and let \(t_{\mathrm{last}}\) be the last model-output timestamp (\texttt{reasoning}, \texttt{text}, or \texttt{tool\_call}). For a step with provider-reported \outputtoks \(O\), we define normalized decode speed as
\[
  s_{\mathrm{norm}} = \frac{O}{t_{\mathrm{last}} - t_{\mathrm{input}}}.
\]
This is an end-to-end trace speed: it includes TTFT, reasoning, visible output emission, and trace logging effects.

For Codex steps with exact reasoning-token accounting, let \(O_{\mathrm{reason}}\) be reasoning tokens, \(O_{\mathrm{visible}}=O-O_{\mathrm{reason}}\) be non-reasoning \outputtoks, \(t_{\mathrm{reason}}\) be the reasoning timestamp, and \(t_{\mathrm{visible}}\) be the last non-reasoning model-output timestamp.  We then estimate pure decode speed and residual TTFT as
\[
\begin{aligned}
  s_{\mathrm{pure\_decode}} &=
    \frac{O_{\mathrm{visible}}}{t_{\mathrm{visible}} - t_{\mathrm{reason}}}, \\
  \hat{\ell}_{\mathrm{pure\_decode}} &=
    \frac{\sum (t_{\mathrm{visible}} - t_{\mathrm{reason}})}
         {\sum O_{\mathrm{visible}}}, \\
  \widehat{\mathrm{TTFT}} &=
    (t_{\mathrm{reason}} - t_{\mathrm{input}})
    - O_{\mathrm{reason}} \hat{\ell}_{\mathrm{pure\_decode}}.
\end{aligned}
\]

\cref{fig:context_decode_speed} plots these timing metrics against the total input context length. \new{The median step is \mediandecodespeedclaude and \mediandecodespeedcodex normalized tokens/s for Claude and Codex, respectively, but the per-step variance is large at every context length.} Claude's binned median stays near 50--54 tokens/s through most of its range and drops only at the longest contexts, reaching roughly 43 tokens/s around 740k input tokens. Codex shows a clearer context-length trend: its binned median falls from about 43 tokens/s around 12k--23k input tokens to about 29 tokens/s near 185k. In the Codex-only panels, median pure decode speed falls from about 74 tokens/s around 12k input tokens to about 55 tokens/s near 185k, while median residual TTFT increases from about 1.5 s to 2.9 s over the same range. Thus, longer context is associated with slower observed generation, especially for Codex, but context length alone does not explain the wide spread; scheduling, model version, output shape, and backend state likely also contribute.

\subsection{Takeaways and Systems Opportunities} 
\textbf{Takeaways}
\begin{itemize}
    \item \textbf{Long context, short output.} A typical step reads on the order of 100k (mostly cached) input tokens but emits only hundreds of \outputtoks.
    \item \textbf{Bimodal input structure.} Steps with short prefixes tend to carry large \appendtoks, whereas when the prefix is long, the \appendtoks are usually incremental.
    \item \textbf{Mixed output attribution.} Claude and GPT-5.5 primarily use resend output as next round input, whereas GPT-5.4 primarily treats output as part of next round's prefix.
    \item \textbf{Noisy decoding speed.} Longer contexts are associated with slower observed generation, especially for Codex, but the variance remains large even at similar context lengths.
\end{itemize}
\textbf{Systems Opportunities}
\begin{itemize}
    \item \textbf{Append-aware prefill optimization.} The bimodal input structure suggests two different optimization targets. Rare long-append steps should be optimized for prefill throughput, while the common short-append steps may need another path for latency optimizations. 
    \item \textbf{Output-cache policy design.} The coexistence of output-cached and output-resend behavior exposes a tradeoff between transferring decode-produced KV entries back to shared storage and re-prefilling prior outputs in the next step. Serving systems should choose this policy based on output length, cache pressure, and KV cache infra.
    \item \textbf{Dedicated TTFT optimization.} The significant residual TTFT for Codex suggests that better autoscaling, request routing, and decode admission control could meaningfully reduce end-to-end latency, especially for this multi-step coding-agent workflows.
\end{itemize}

\section{Tool Calls}
In this section, we analyze coding agents' tool-calling behavior, including tool call counts, latency, and human interactions.

\subsection{Tool Call Count Distribution}
\begin{figure}[t]
\centering
\includegraphics[width=\columnwidth]{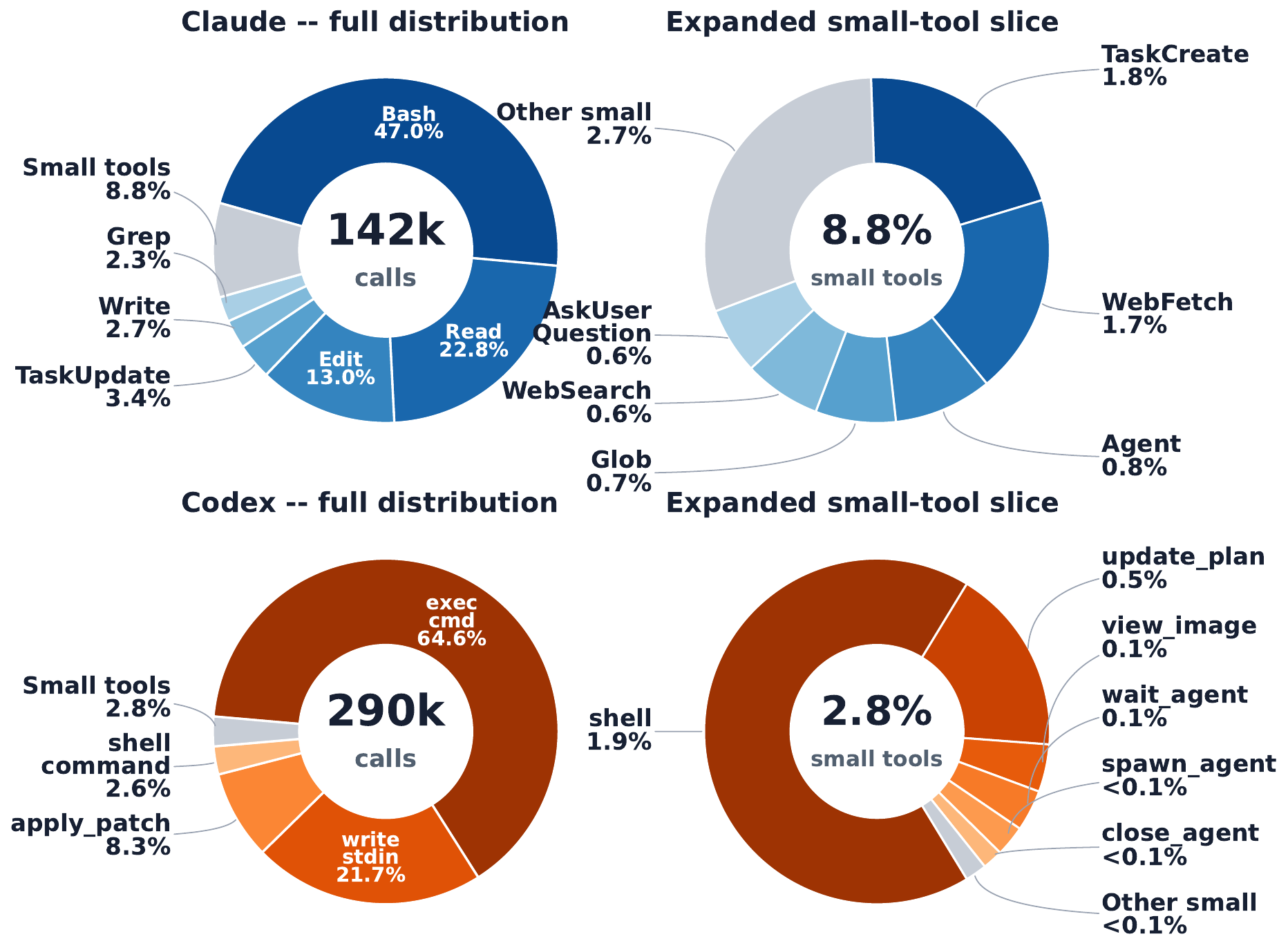}
\caption{Tool call count distribution for Claude and Codex. Command execution dominates both, followed by file operations such as read and edit.}
\label{fig:tool_call_counts}
\end{figure}

We first investigate tool call counts. In the trace, we observe \toolcallkindsclaude different tools for Claude and \toolcallkindscodex for Codex. Tool popularity is shown in \cref{fig:tool_call_counts}. The most common tool is command execution for both models, followed by file operations such as Read and Edit. For Claude, the top three tools account for over \toolcalltoppercentage of calls, while for Codex they account for \toolcalltoppercentagecodex.

\begin{figure}[t]
\centering
\includegraphics[width=\columnwidth]{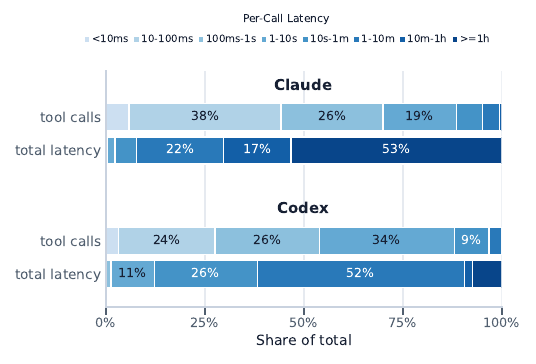}
\caption{Tool-call latency bins by call share and by total-latency share, split by provider.}
\label{fig:tool_latency_weighted_bins}
\end{figure}

\begin{figure*}[t]
\centering
\includegraphics[width=\textwidth]{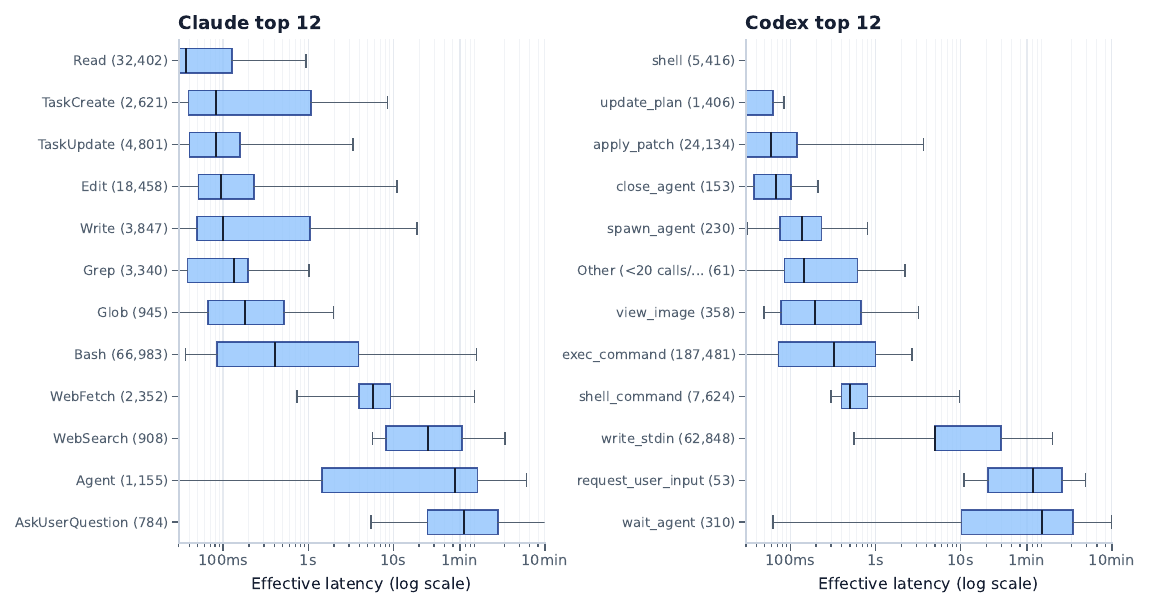}
\caption{Effective tool-call latency for the top 12 tools by call count in each provider. Boxes show the interquartile range, center lines show medians, and whiskers show p5--p95.}
\label{fig:tool_latency_by_tool_top12}
\end{figure*}

\subsection{Tool Call Latency Distribution}
Next, we analyze tool call latency. \Cref{fig:tool_latency_by_tool_top12} shows the latency distribution for the top 12 most common tools for both Claude and Codex. Overall, average tool-call latency is 16.8s, but latency depends heavily on tool type. For example, file operations such as Read and Edit are mostly short, on the order of milliseconds to seconds, while \texttt{Agent} and \texttt{AskUserQuestion} can take minutes to finish. Even within each tool type, however, latency varies significantly. For example, Claude's \texttt{Bash} tool ranges from milliseconds to minutes, although its median is below one second.

Tool-call latency has a strong long tail. \Cref{fig:tool_latency_weighted_bins} compares, for each latency bin, its share of tool calls with its share of total tool-call time. For Claude, calls under 1\,s account for 70\% of calls but less than 1\% of total tool time, while calls longer than 1\,min are only 4.9\% of calls but contribute 92\% of total tool time. Codex is less extreme but still dominated by slow calls: calls under 10\,s account for 88\% of calls but only 12\% of time, while calls longer than 1\,min are 3.1\% of calls and 61\% of time.

\subsection{Tool Call Overhead}
Codex exposes two timing views for most tool calls: the trace-observed end-to-end span from tool emission to tool result, and the runner-reported internal execution time. For each call we defining overhead as $R=\max(0, T_{\mathrm{e2e}}-T_{\mathrm{int}})$. This captures time around a tool call that is not reported as internal execution, which can include permission approval, runtime scheduling, shell startup, output propagation, client-side bookkeeping, etc.

\begin{table}[t]
\centering
\caption{Codex end-to-end and internal tool latency, with positive residual statistics.}
\label{tab:codex_tool_e2e_internal}
\footnotesize 
\setlength{\tabcolsep}{1.3pt}
\renewcommand{\arraystretch}{1.10}
\begin{tabular*}{\columnwidth}{@{\extracolsep{\fill}}l r r r r r r@{}}
\toprule
\textbf{Tool} & \textbf{Calls} & \textbf{E2E} & \textbf{Int.} & \textbf{Res.} & \textbf{Avg} & \textbf{P50/90/99} \\
\midrule
All timed & 253k & 418.1h & 341.5h & 77.8h & 1.11s & 0.13/0.24/10.0s \\
\texttt{exec\_command} & 184k & 97.8h & 33.8h & 64.1h & 1.26s & 0.16/0.27/8.6s \\
\texttt{write\_stdin} & 61k & 314.6h & 303.7h & 11.7h & 0.69s & 0.01/0.10/14.9s \\
\texttt{shell\_command} & 6.1k & 5.3h & 3.8h & 1.8h & 1.08s & 0.04/0.16/4.0s \\
\texttt{apply\_patch} & 2.5k & 0.5h & 0.3h & 0.21h & 0.30s & 0.01/0.41/3.9s \\
\bottomrule
\end{tabular*}
\end{table}

\Cref{tab:codex_tool_e2e_internal} reports this overhead for the 253,391 Codex tool calls with valid end-to-end and internal timings, covering 87.3\% of Codex tool calls. These calls account for 418.1h end-to-end and 341.5h of runner-reported internal execution, leaving 77.8h of residual (18.6\%). The residual is dominated by \texttt{exec\_command}, which contributes 64.1h. The average residual is 1.11s, and the p50/p90 remain small (0.13s/0.24s), but the p99 reaches 10.0s. Thus, aggregate overhead comes from many small gaps plus a long tail. Our hypothesis is that bash commands more likely require human or auto approval, which enlarges the p99 latency. Claude does not expose comparable internal timing coverage, so we do not make the same quantitative decomposition for Claude.

\subsection{Takeaways and System Opportunities}
\textbf{Takeaways}
\begin{itemize}
    \item \textbf{Heavily skewed tool call counts and latency}. A small number of tools dominate call counts, and a small number of slow calls dominate total tool-call time.
    \item \textbf{Per-type latency variation} Tool type can guide latency prediction, but there is still significant variation within each tool type.
    \item \textbf{Considerable overhead} For Codex, observed end-to-end tool latency can substantially exceed runner-reported internal execution time, indicating non-execution overheads.
\end{itemize}
\textbf{System Opportunities}
\begin{itemize}
    \item \textbf{Semantic-based latency prediction } When predicting tool-call latency for \pfc management, tool type alone is insufficient. The large variation within each tool type suggests using semantic information about the requested operation, together with recent latency history. \cite{tiwari2026cachewiseunderstandingworkloadsoptimizing}
    \item \textbf{Low overhead tool calling} Developing low-latency auto-approval and reducing tool-call framework overhead can reduce tool call overhead, which is a non-negligible part of tool call latency and overall session latency.
\end{itemize}

\section{Prefix Cache}

\begin{table}[t]
\centering
\caption{Token-weighted prefix cache hit rate by provider and step trigger.}
\label{tab:prefix_cache_hit_rate}
\small
\setlength{\tabcolsep}{6pt}
\renewcommand{\arraystretch}{1.15}
\begin{tabular}{l r r r}
\toprule
\textbf{Metric} & \textbf{Claude} & \textbf{Codex} & \textbf{Total} \\
\midrule
Prefix cache-hit rate & 95.8\% & 95.7\% & 95.7\% \\
Prefix hit rate (user-initiated) & 86.9\% & 78.2\% & 84.4\% \\
Prefix hit rate (tool-result) & 97.9\% & 97.2\% & 97.5\% \\
\bottomrule
\end{tabular}
\end{table}

\begin{figure*}[t]
\centering
\includegraphics[width=\textwidth]{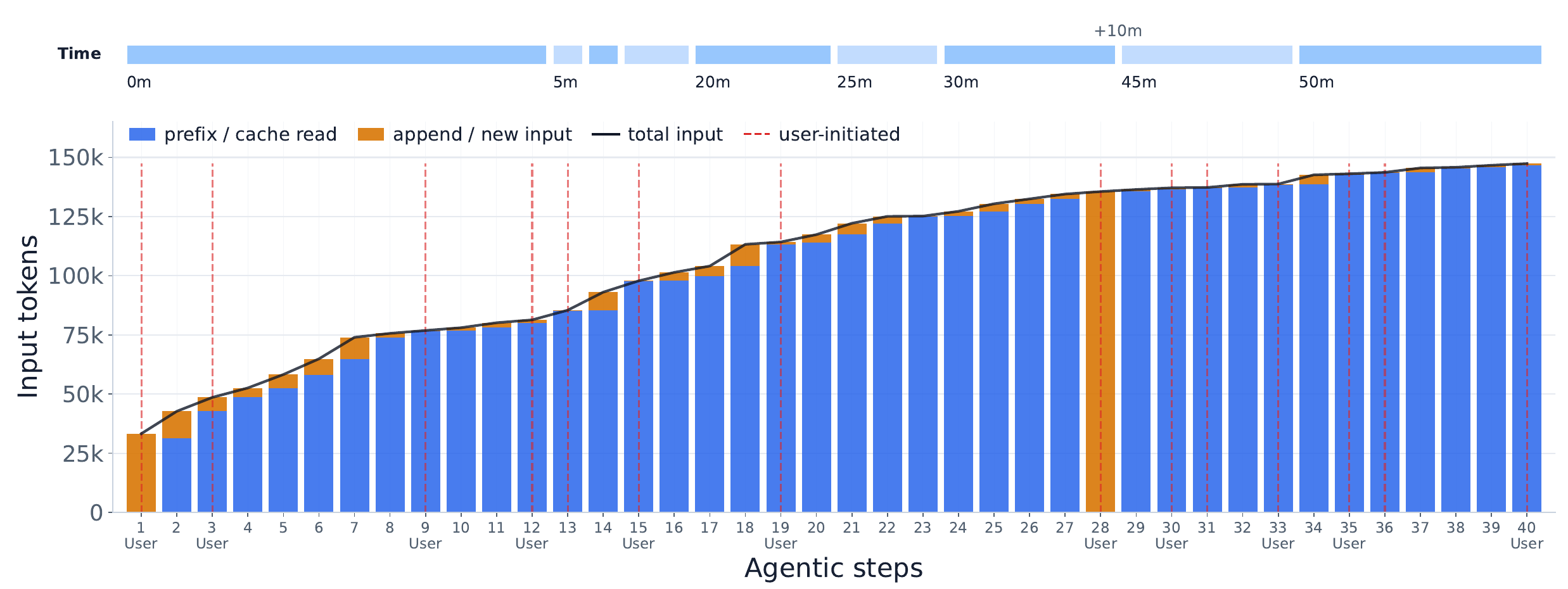}
\caption{Example session progression over the first 40 agentic steps. Bars decompose each step input into \prefixtoks and \appendtoks; the black line shows total input, red dashed lines mark user-initiated steps, and the top strip shows elapsed time.}
\label{fig:session_progress_example}
\end{figure*}

\begin{figure*}[t]
\centering
\includegraphics[width=\textwidth]{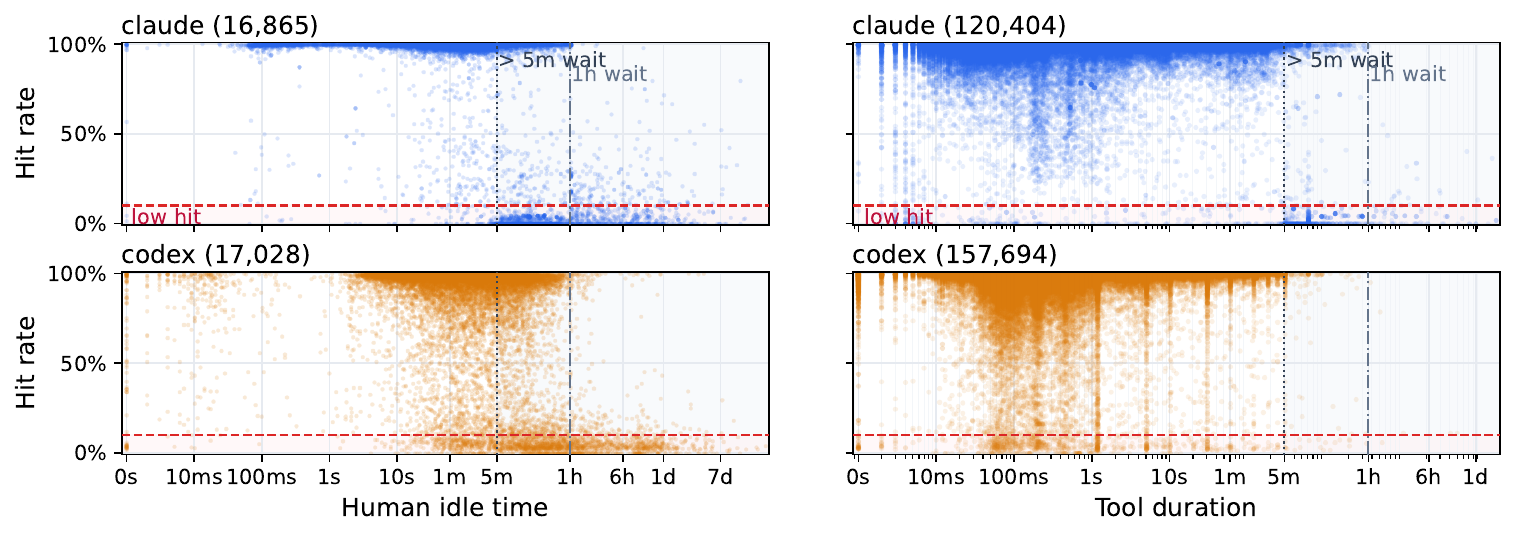}
\caption{\Pfc hit rate versus the idle time preceding a step (log $x$-axis), for Claude (top) and Codex (bottom).}
\label{fig:prefix_cache_hit_rate_by_idle_time}
\end{figure*}

\begin{table}[t]
\centering
\caption{Fresh prefill versus total append tokens, by provider and step trigger. \emph{Fresh} tokens are the per-step context growth minus the prior step's output tokens; \emph{prefill amplification} is append tokens${}/{}$fresh.}
\label{tab:redundant_prefill}
\small
\setlength{\tabcolsep}{6pt}
\renewcommand{\arraystretch}{1.15}
\begin{tabular}{l r r r}
\toprule
\textbf{Metric} & \textbf{Claude} & \textbf{Codex} & \textbf{Total} \\
\midrule
\multicolumn{4}{@{}l}{\emph{Overall}} \\
\quad Total append tokens & 1,154.0\,M & 1,115.6\,M & 2,269.6\,M \\
\quad Total fresh tokens & 142.1\,M & 288.1\,M & 430.2\,M \\
\quad Fresh \% of append & 12.3\% & 25.8\% & 19.0\% \\
\quad Prefill amplification & 8.1$\times$ & 3.9$\times$ & 5.3$\times$ \\
\addlinespace
\multicolumn{4}{@{}l}{\emph{User-initiated}} \\
\quad Total append tokens & 672.1\,M & 454.7\,M & 1,126.8\,M \\
\quad Total fresh tokens & 11.4\,M & 20.4\,M & 31.8\,M \\
\quad Fresh \% of append & 1.7\% & 4.5\% & 2.8\% \\
\quad Prefill amplification & 58.8$\times$ & 22.3$\times$ & 35.4$\times$ \\
\addlinespace
\multicolumn{4}{@{}l}{\emph{Tool-result}} \\
\quad Total append tokens & 481.8\,M & 661.0\,M & 1,142.8\,M \\
\quad Total fresh tokens & 130.7\,M & 267.7\,M & 398.4\,M \\
\quad Fresh \% of append & 27.1\% & 40.5\% & 34.9\% \\
\quad Prefill amplification & 3.7$\times$ & 2.5$\times$ & 2.9$\times$ \\
\bottomrule
\end{tabular}
\end{table}

\begin{figure}[t]
\centering
\includegraphics[width=\columnwidth]{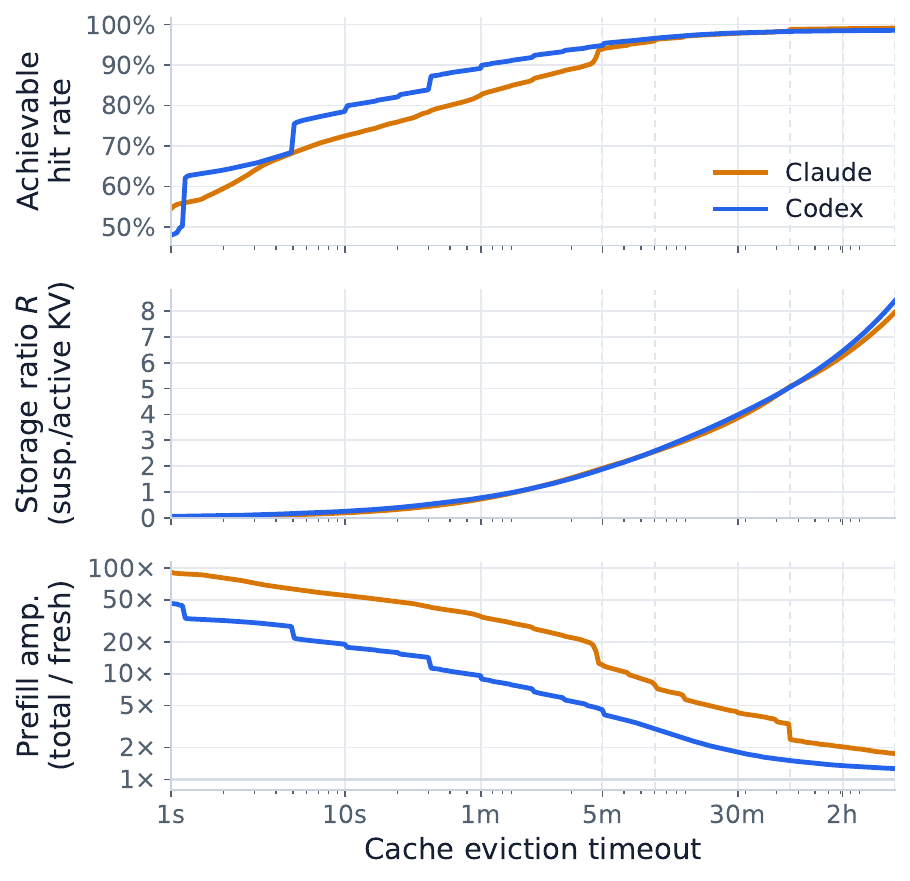}
\caption{\Pfc eviction trade-off versus the eviction timeout (shared log $x$-axis), for
Claude and Codex: achievable hit rate (top), storage ratio $R$ of suspended to active KV (middle),
and prefill amplification (bottom).}
\label{fig:eviction_tradeoff}
\end{figure}

In this section, we examine a key component of LLM serving for coding agents: the \pfc. We analyze its hit rate and how it relates to idle time. Then we characterize the gap to the optimal, the trade-off between KV storage space and the eviction time of cached prefixes, and the consumer-visible cost of human thinking time.

\label{sec:prefixcache}
\subsection{Per-Step Prefix Cache Hit Rate}
To begin with, we break down the cache hit rate by step type. \Cref{tab:prefix_cache_hit_rate} shows that prefix caching is consistently high overall: both Claude and Codex serve about 96\% of prompt tokens from the \pfc. The main misses are user-initiated steps, where human thinking time can make the idle time long enough to trigger eviction; tool-result steps remain near-perfect because they usually resume shortly.

In \cref{fig:session_progress_example}, we further illustrate the \pfc hit rate over an example session. Most steps are initiated close together in time, so the \pfc retains the context. At step 28, a cache miss occurs due to 10 minutes of human inactivity, causing a large fresh prefill.

\subsection{Prefix Cache Eviction Time}
Next, we dive into the \pfc eviction time. We measure the idle time between steps and plot the corresponding \pfc hit rate in \cref{fig:prefix_cache_hit_rate_by_idle_time}. For both Claude and Codex, long gaps are significantly more common for user-initiated steps than for tool-result steps. When the gap is larger than 5 minutes, low-hit-rate steps begin to appear, and after 1 hour, almost all steps miss the cache. Tool gaps, especially for Codex, rarely exceed 5 minutes, since Codex moves long-running tasks into the background.

\subsection{Gap to Optimal and Redundant Prefill}

While the overall cache hit ratio is high, there is still a gap to optimal. Specifically, the upper bound of the \pfc hit rate is determined by the user prompts and tool results, since these are truly fresh tokens the system has never seen before. To measure them precisely, we take the total context growth and subtract the \outputtoks. \Cref{tab:redundant_prefill} reports these fresh tokens against the total \appendtoks. Across all steps, fresh tokens are only 19.0\% of \appendtoks (12.3\% for Claude, 25.8\% for Codex), meaning the remaining $\sim$81\% of prefill can, in principle, be served from cache, marking the gap to optimal. The split is sharply step-dependent: user-initiated steps are almost entirely re-sent context (fresh is just 1.7\% of their \appendtoks for Claude and 4.5\% for Codex), due to their lower cache hit rate, whereas tool-result steps carry more of the genuinely new content (27.1\% and 40.5\%, respectively).

We further compute the \emph{prefill amplification factor}, the ratio of total prefilled tokens to the irreducible fresh tokens---equivalently $1/(\text{fresh \% of append})$---i.e., how many times more tokens are prefilled than a perfect, eviction-free cache would require. The deployed cache amplifies prefill by $5.3\times$ overall ($8.1\times$ for Claude, $3.9\times$ for Codex), flagging an optimization opportunity.

\subsection{Storage Trade-off}
We next examine the trade-off between the \pfc storage space and the hit rate. Based on the observed human thinking time, tool latency, and LLM generation time, we vary the cache eviction time and compute the corresponding hit rate and storage space. With a global request batch $B$, the fraction of request that is actively decoding is approximately $B \cdot T_{generation} / (\hat{T}_{human}+\hat{T}_{tool}+T_{generation})$, where $T_{generation}$ is the average LLM generation time, and $\hat{T}_{human}$ and $\hat{T}_{tool}$ are the average human thinking time and tool latency, capped by the eviction time. This indicates that the ratio of the suspended requests' KV cache storage to that of the active decoding requests is approximately
$$
    R = \frac{\hat{T}_{human}+\hat{T}_{tool}}{T_{generation}}.
$$

We sweep the eviction time and derive the corresponding achievable hit rate and prefill amplification under an idealized rule: a step is a complete cache miss when its preceding idle gap exceeds the eviction time, and a full hit otherwise. \Cref{fig:eviction_tradeoff} plots all three quantities against a shared eviction-timeout axis. Raising the timeout from 1\,min to 1\,h lifts the achievable hit rate from 85.4\% to 98.6\%, but increases the storage ratio from $R=0.74$ to $R=5.07$ ($\sim$7$\times$ more suspended KV). Most of the gain is cheap: by 5\,min the hit rate is already $\sim$94\% at $R\approx1.9$, and the remaining push to 1\,h costs $\sim$2.7$\times$ more storage for only $\sim$4 more points of hit rate.

The prefill amplification panel (bottom) tells a similar story. An idealized cache that never evicts would prefill only fresh tokens, so its amplification \emph{floors at $1\times$}; tightening the eviction time re-prefills evicted context and inflates it (merged $18.9\times$ at 1\,min, $7.4\times$ at 5\,min, $1.8\times$ at 1\,h).

\subsection{Cost of Human Thinking Time}
The eviction sweep above is a system/operator view: longer retention improves hit rate but consumes more KV storage. We now analyze the same \pfc behavior from a consumer perspective. For a user, ``thinking'' between requests can become a direct cost: if the session prefix expires during the pause, the next user-initiated step pays the fresh-input price to prefill context that was already present before the pause.

We estimate an upper bound on this consumer-side cost by asking how much append prefill could be avoided if user-initiated steps retained their \pfc across human thinking time. For each user-initiated step $S$ with a predecessor step $P$ in the same session, we keep the total input length unchanged but cap the \appendtoks at the net context growth, $\max(0, L_S-L_P)$, where $L$ is \prefixtoks plus \appendtoks. If the observed \appendtoks is already smaller than this value, we leave it unchanged. Tool-result steps and session-first steps are also unchanged. This assumes shifted tokens can be served from cache at the cache-read rate, so the resulting savings are an upper bound rather than an achievable policy guarantee. \Cref{tab:human_thinking_cost} reports the resulting token and cost reductions.

\begin{table}[t]
\centering
\caption{Upper-bound append-token and cost savings from eliminating user-thinking-induced
\pfc misses.}
\label{tab:human_thinking_cost}

\setlength{\tabcolsep}{2pt}
\renewcommand{\arraystretch}{1.15}
\resizebox{\columnwidth}{!}{%
\begin{tabular}{l r r r}
\toprule
\textbf{Metric} & \textbf{Claude} & \textbf{Codex} & \textbf{Total} \\
\midrule
User steps w/ predecessor & 16,927 & 17,033 & 33,960 \\
Observed append & 1.19\,B & 1.15\,B & 2.34\,B \\
Append after retained cache & 541.9\,M & 721.7\,M & 1.26\,B \\
Append reduction & 648.1\,M (54.5\%) & 423.9\,M (37.0\%) & 1.07\,B (45.9\%) \\
Observed total cost & \$22,654 & \$17,777 & \$40,431 \\
Cost after retained cache & \$18,973 & \$16,269 & \$35,242 \\
Cost reduction & \$3,680 (16.2\%) & \$1,508 (8.5\%) & \$5,189 (12.8\%) \\
Avg saved / reduced step & \$0.263 & \$0.116 & \$0.192 \\
\bottomrule
\end{tabular}
}
\end{table}

\Cref{tab:human_thinking_cost} shows that this upper-bound estimate reduces append prefill by 1.07\,B tokens overall, or 45.9\% of observed \appendtoks. At current list prices, this corresponds to a final cost reduction of \$5,189 (12.8\%) over priced rounds. The reduction is larger for Claude in both token and dollar terms due to its longer average context length.

\subsection{Takeaways and System Opportunities}
\textbf{Takeaways}
\begin{itemize}
    \item \textbf{User-initiated steps miss the cache.} Overall \pfc hit rate is high, but user-initiated steps are more likely to miss the cache due to longer idle time.
    \item \textbf{Idle time driven misses.} Cache misses mainly increase when idle time is larger than 5\, min and almost never hit when past 1\,h.
    \item \textbf{Large gap to optimal.} Only 19\% of \appendtoks are fresh, indicating a large gap to optimal and significant redundant prefill, and a trade-off exists between hit rate and storage space.
    \item \textbf{Human thinking time is costly.} In our retained-cache upper-bound estimate, cache misses after human thinking time account for 45.9\% of \appendtoks and 12.8\% of total priced cost.
\end{itemize}
\textbf{System Opportunities}
\begin{itemize}
    \item \textbf{Step type driven eviction.} Choosing the right eviction time depends on the ratio of cache storage cost to prefill cost. Advances in KV compression and reductions in storage cost can help reduce the prefill workload.
    \item \textbf{Prediction based prefetching.} Gap-duration prediction and KV cache prefetching can evict the cache more aggressively while still providing similar TTFT for the user.
    \item \textbf{Keeping the cache alive.} Agent harnesses can periodically refresh the cache during long human-thinking gaps to avoid eviction and reduce prefill cost, or reduce the frequency of user interventions through better automation.
\end{itemize}

\section{Related Work}

\subsection{Real-world coding-agent traces and developer usage}

Some existing work studies coding agents and AI-assisted programming in
real use. CacheWise~\cite{tiwari2026cachewiseunderstandingworkloadsoptimizing}
is a companion systems study from the same broad research direction: it uses
CATraces, a real Claude Code trace, to characterize closed-loop coding-agent
requests and to design a vLLM KV-cache layer with prefix-aware scheduling and
predictive eviction from tool-call metadata. \sys{} builds on this line but analyzes a much
broader cross-provider trace spanning Claude Code and Codex with several orders of magnitude larger token counts. 
SWE-chat~\cite{baumann2026swechatcodingagentinteractions} collects thousands of
real coding-agent sessions from open-source developers and analyzes behavioral
outcomes such as vibe-coding, code survival, and security. AgentPack~\cite{zi2026agentpackdatasetcodechanges}
and The Rise of AI Teammates in Software Engineering~\cite{li2025riseaiteammatessoftware}
collect large numbers of agent-authored code edits or pull requests from
GitHub. These works analyze
code-change outcomes but not the interaction timeline or serving system that produced them.
Programming by Chat~\cite{tang2026programmingchatlargescalebehavioral},
Reading Between the Lines~\cite{mozannar2024readinglinesmodelinguser},
the GitHub Copilot productivity study~\cite{ziegler2022productivityassessmentneuralcode}, the
professional-developer qualitative study~\cite{huang2025professionalsoftwaredevelopersdont},
and the Cursor adoption study~\cite{he2026speedcostqualitycursor} similarly
characterize how developers use AI coding tools, ranging from IDE chat logs and
surveys to repository-level productivity and technical-debt outcomes. \sys{} is mainly system-focused, investigating the system implications of agentic workloads.

\subsection{Production LLM serving traces and conversation datasets}

Large-scale LLM serving traces have driven many top-tier systems papers, but
their workloads are mostly chat, API, or code-completion traffic. Mooncake~\cite{qin2025mooncakekvcachecentricdisaggregatedarchitecture} releases a replayed production trace with arrival times, token counts, and remapped block hashes for the Kimi serving stack. Splitwise~\cite{patel2024splitwiseefficientgenerativellm} characterizes Azure ``coding'' and ``conversation'' traces, where the coding trace is closer to Copilot-style completion than multi-step agent execution. BurstGPT~\cite{wang2025burstgptrealworldworkloaddataset}, DynamoLLM~\cite{stojkovic2024dynamollmdesigningllminference}, ServeGen~\cite{xiang2026servegenworkloadcharacterizationgeneration}, KVCache Cache in the Wild~\cite{wang2026kvcachecachewildcharacterizing}, POLCA~\cite{patel2023polcapoweroversubscriptionllm}, and BatchLLM~\cite{zheng2026batchllmoptimizinglargebatched} study burstiness, energy, request generation, multi-turn shared-prefix reuse, power oversubscription, and global prefix sharing in production or production-like LLM traffic. A second family of datasets, including LMSYS-Chat 1M~\cite{zheng2024lmsyschat1mlargescalerealworldllm}, WildChat~\cite{zhao2024wildchat1mchatgptinteraction}, MT-Bench~\cite{zheng2023judgingllmasajudgemtbenchchatbot}, Chatbot Arena~\cite{chiang2024chatbotarenaopenplatform}, OpenAssistant~\cite{kopf2023openassistantconversationsdemocratizing}, WildBench~\cite{lin2024wildbenchbenchmarkingllmschallenging}, and Clio~\cite{tamkin2024clioprivacypreservinginsightsrealworld}, captures real conversational use or derives benchmarks from user conversations. These traces are useful for trace-driven serving research, but they do not expose the session/request/step structure of coding agents: long accumulated context, frequent tool-triggered model invocations, tool latency, and human-paced gaps between requests.

\sys{} fills this missing
workload category by measuring real coding-agent traffic at the systems detail
needed for serving decisions.

\subsection{Coding-agent benchmarks and tool-use evaluation}

Software-engineering benchmarks evaluate whether agents can solve tasks.
SWE-bench~\cite{jimenez2024swebenchlanguagemodelsresolve} and its variants,
including SWE-bench Multimodal~\cite{yang2024swebenchmultimodalaisystems},
Multi-SWE-bench~\cite{zan2025multiswebenchmultilingualbenchmarkissue},
SWE-Gym~\cite{pan2025trainingsoftwareengineeringagents}, and SWE-Lancer~\cite{miserendino2025swelancerfrontierllmsearn},
use realistic issue-resolution tasks. Terminal-Bench~\cite{merrill2026terminalbenchbenchmarkingagentshard}
evaluates agents in containerized terminal environments. LiveCodeBench~\cite{jain2024livecodebenchholisticcontaminationfree},
BigCodeBench~\cite{zhuo2025bigcodebenchbenchmarkingcodegeneration}, HumanEval~\cite{chen2021evaluatinglargelanguagemodels},
MBPP~\cite{austin2021programsynthesislargelanguage}, DS-1000~\cite{lai2022ds1000naturalreliablebenchmark},
APPS~\cite{hendrycks2021measuringcodingchallengecompetence}, RepoBench~\cite{liu2023repobenchbenchmarkingrepositorylevelcode},
ClassEval~\cite{du2023classevalmanuallycraftedbenchmarkevaluating}, CrossCodeEval~\cite{ding2023crosscodeevaldiversemultilingualbenchmark},
Commit0~\cite{zhao2024commit0librarygenerationscratch}, MLE-bench~\cite{chan2025mlebenchevaluatingmachinelearning},
and Prompting LLMs to Tackle the Full Software Development Lifecycle~\cite{li2024promptinglargelanguagemodels} cover competitive
programming, library use, data science, repository context, cross-file
completion, whole-library generation, ML engineering, and full software
development workflows. Agent frameworks and tool-use benchmarks provide the
execution substrate for these tasks: SWE-agent~\cite{yang2024sweagentagentcomputerinterfacesenable}
and OpenHands~\cite{wang2025openhandsopenplatformai} implement tool-using
software-engineering agents; Agentless~\cite{xia2024agentlessdemystifyingllmbasedsoftware}
and AutoCodeRover~\cite{zhang2024autocoderoverautonomousprogramimprovement}
explore more structured repair pipelines; ReAct~\cite{yao2023reactsynergizingreasoningacting}
and Toolformer~\cite{schick2023toolformerlanguagemodelsteach} define
influential reasoning-and-action and tool-learning paradigms; tau-bench~\cite{yao2024taubenchbenchmarktoolagentuserinteraction},
tau2-bench~\cite{barres2025tau2benchevaluatingconversationalagents}, ToolLLM~\cite{qin2023toolllmfacilitatinglargelanguage},
Gorilla~\cite{patil2023gorillalargelanguagemodel}, AgentBench~\cite{liu2025agentbenchevaluatingllmsagents},
WebArena~\cite{zhou2024webarenarealisticwebenvironment}, GAIA~\cite{mialon2023gaiabenchmarkgeneralai},
and OSWorld~\cite{xie2024osworldbenchmarkingmultimodalagents} test general tool
use, web interaction, computer use, and multi-step assistance.

These benchmarks and frameworks are essential for capability evaluation, but they are
intentionally task-centric: a benchmark instance is usually a single isolated
request, often with constrained tooling and limited session history. \sys{}
instead measures whole day-to-day sessions, capturing accumulated context,
human delays, repeated tool calls, cache retention, and serving cost that
benchmark replay cannot faithfully represent.

\subsection{LLM serving systems and agent-aware scheduling}

Modern LLM serving systems optimize batching, memory management, disaggregation,
and scheduling. Orca~\cite{yu2022orca} introduced iteration-level batching
for variable-length generation. vLLM/PagedAttention~\cite{kwon2023efficientmemorymanagementlarge}
manages KV memory in non-contiguous pages and enables prefix sharing.
Sarathi-Serve~\cite{agrawal2024tamingthroughputlatencytradeoffllm}, DistServe~\cite{zhong2024distservedisaggregatingprefilldecoding},
Splitwise~\cite{patel2024splitwiseefficientgenerativellm}, Inference without
Interference~\cite{hu2024inferenceinterferencedisaggregatellm}, and
DeepSpeed-FastGen~\cite{holmes2024deepspeedfastgenhighthroughputtextgeneration}
target the prefill/decode throughput-latency trade-off through chunked prefills,
phase splitting, or disaggregated execution. FlexGen~\cite{sheng2023flexgenhighthroughputgenerativeinference}
studies throughput-oriented offloading, AlpaServe~\cite{li2023alpaservestatisticalmultiplexingmodel} investigates
statistical multiplexing, Fast Distributed Inference Serving~\cite{wu2024fastdistributedinferenceserving} performs
token-level preemptive scheduling, Llumnix~\cite{sun2024llumnixdynamicschedulinglarge} enables
live migration, LoongServe~\cite{wu2024loongserveefficientlyservinglongcontext} proposes
elastic sequence parallelism, and NanoFlow~\cite{zhu2025nanoflowoptimallargelanguage} presents
Intra-device parallelism, these work greatly enhanced the throughput and SLO attainment for large deployments. Agent- and program-aware systems extend this direction:
SGLang~\cite{zheng2024sglangefficientexecutionstructured} support RadixAttention for prefix reuse, Parrot~\cite{lin2024parrotefficientservingllmbased}
schedules application dataflow via semantic variables, Autellix~\cite{luo2025autellixefficientservingengine}
treats agent executions as dynamic programs or DAGs, and VibeServe~\cite{kamahori2026vibeserveaiagentsbuild}
uses agents to tailor serving stacks for specific use cases.

These systems provide the mechanisms to serve coding-agent traffic, and
several already recognize that agent programs are not independent stateless
requests. \sys{} provides a standard benchmark to guide the evolve of serving engines 
for real coding agents.

\subsection{Prefix-cache and KV-state management}

KV reuse is central to efficient multi-turn and agentic serving. SGLang's
RadixAttention~\cite{zheng2024sglangefficientexecutionstructured}, Mooncake~\cite{qin2025mooncakekvcachecentricdisaggregatedarchitecture},
Preble~\cite{srivatsa2024prebleefficientdistributedprompt}, Pensieve~\cite{yu2024statefullargelanguagemodel},
ChunkAttention~\cite{ye2024chunkattentionefficientselfattentionprefixaware}, Hydragen~\cite{juravsky2024hydragenhighthroughputllminference},
Prompt Cache~\cite{gim2024promptcachemodularattention}, CacheBlend~\cite{yao2025cacheblendfastlargelanguage},
CacheGen~\cite{liu2024cachegenkvcachecompression}, EPIC~\cite{hu2025epicefficientpositionindependentcaching},
and LMCache~\cite{liu2025lmcacheefficientkvcache} study different ways to reuse,
place, retrieve, compress, or recompute KV state across requests and prompt
fragments. CachedAttention/AttentionStore~\cite{gao2024costefficientlargelanguagemodel}
and InferCept~\cite{abhyankar2024inferceptefficientinterceptsupport} are
especially close to interactive or augmented LLM settings: they consider saving
session KV across idle periods or choosing among GPU retention, CPU swapping,
and recomputation when generation pauses for tools or humans. DualPath~\cite{wu2026dualpathbreakingstoragebandwidth}
observes that agentic workloads can shift the bottleneck to loading large KV
state from external storage, while Continuum~\cite{li2026continuumefficientrobustmultiturn}
selects a TTL for finished-request KV using empirical tool-call distributions
and scheduling costs.

\sys{} pinpoints the noisy tool latency and human thinking time of the coding trace, which requires further research on how to manage the \pfc.

\subsection{Long-context attention and KV compression}

Another line of work reduces the cost of long contexts by changing attention or
compressing KV state. StreamingLLM~\cite{xiao2024efficientstreaminglanguagemodels},
InfLLM~\cite{xiao2024infllmtrainingfreelongcontextextrapolation}, Native Sparse
Attention~\cite{yuan2025nativesparseattentionhardwarealigned}, MoBA~\cite{lu2025mobamixtureblockattention},
MInference~\cite{jiang2024minference10acceleratingprefilling}, and Quest~\cite{tang2024questqueryawaresparsityefficient}
reduce attention cost through locality and block-level or page-level importance estimation and sparse selection. H2O~\cite{zhang2023h2oheavyhitteroracleefficient},
Scissorhands~\cite{liu2023scissorhandsexploitingpersistenceimportance}, SnapKV~\cite{li2024snapkvllmknowslooking},
PyramidKV~\cite{cai2025pyramidkvdynamickvcache}, DuoAttention~\cite{xiao2024duoattentionefficientlongcontextllm},
KIVI~\cite{liu2024kivituningfreeasymmetric2bit}, and KVQuant~\cite{hooper2025kvquant10millioncontext}
reduce KV footprint through heavy-hitter retention, prompt-token selection,
layer/head-aware budgets, quantization, or mixed full/streaming-head caches.

These techniques are often evaluated on long documents or long-context
benchmarks, but \sys{} demonstrate that the same pressure appears in coding agents because sessions
replay very large prefixes while appending only small tool results or user
turns.

\section{Limitations and Future Work}

Our study has several limitations that also point to future work. First, our trace is drawn from our own day-to-day use of coding agents. Although it spans multiple developers, agents, and model versions, the results may not fully generalize to other organizations, development workflows, or agent domains. Second, our analysis is limited to externally visible logs, including prompts, model responses, tool calls, timestamps, and token metadata. Without internal telemetry from service providers, we can identify workload patterns but can only infer the mechanisms behind them, such as scheduling policies, cache behavior, or backend routing. Third, agent workloads are expanding beyond coding to broader computer-use and automation tasks. Future trace studies should extend this methodology to other agent domains and compare how their context growth, tool use, latency, and cache behavior differ from coding-agent workloads.

\section{Conclusion}
\label{sec:conclusion}

In this paper, we present a trace-based analysis of coding-agent workloads. Notably, we find that these workloads feature long autonomous loops, long contexts paired with short outputs, strongly long-tailed tool calls, and high yet imperfect prefix-cache hit rates. Taken together, these observations point to serving opportunities, including denser or lower-overhead tool calling, append-length-specific prefill optimization, and better KV-cache eviction or prefetching around human-paced gaps. We hope this work provides a useful reference for future system research on coding agents and, more broadly, on LLM-based agents, and we look forward to community efforts that extend this trace-based analysis methodology to larger-scale and more diverse agent workloads.

\bibliographystyle{plain}
\bibliography{_reference}

\end{document}